%% file: arxiv_main.tex
\documentclass[11pt]{article}
\pdfoutput=1
\usepackage[T1]{fontenc}
\usepackage{lmodern} \normalfont %
\usepackage{anyfontsize}
\DeclareFontShape{T1}{lmr}{bx}{sc} { <-> ssub * cmr/bx/sc }{}
\DeclareFontShape{T1}{lmr}{m}{scit}{ <-> ssub * cmr/m/sc }{}
\DeclareFontShape{T1}{lmr}{bx}{scit}{ <-> ssub * cmr/bx/sc }{}

\usepackage{etoolbox}
\usepackage{url}            %

\usepackage[round]{natbib}
\usepackage{booktabs}       %
\usepackage{amsfonts}       %
\usepackage{nicefrac}
\usepackage{microtype}
\usepackage{graphicx}
\usepackage{booktabs} %
\usepackage{subcaption}
\usepackage{lipsum}
\usepackage{tkz-fct}
\usetikzlibrary{patterns}
\usepackage{makecell}
\usepackage{arydshln}
\usepackage{amssymb}
\usepackage{amsmath,amsfonts,bm}
\usepackage{mathtools,enumitem}

\usepackage{amsthm}
\usepackage{algorithm}
\usepackage{algorithmicx}
\usepackage[noend]{algpseudocode}
\usepackage{algorithm}
\usepackage{algorithmicx}
\usepackage[noend]{algpseudocode}

\usepackage{titlesec}
\titlespacing{\section}{0pt}{\parskip}{0pt}
\titlespacing{\subsection}{0pt}{\parskip}{0pt}
\titlespacing{\subsubsection}{0pt}{\parskip}{0pt}
\usepackage[letterpaper,margin=0.85in]{geometry}
\definecolor{mylinkcolor}{rgb}{0.7, 0.0, 0.0}
\definecolor{mycitecolor}{rgb}{0.5, 0.0, 0.13}
\definecolor{myurlcolor}{rgb}{0.7, 0.0, 0.7}

\usepackage{hyperref}
\hypersetup{
    linktocpage=true,
    colorlinks=true,
    linkcolor=mylinkcolor,
    citecolor=mycitecolor,
    urlcolor=myurlcolor,
}

\usepackage[capitalize,noabbrev]{cleveref}

\input{math_commands}

\newtheorem{theorem}{Theorem}[section]
\newtheorem{proposition}[theorem]{Proposition}
\newtheorem{definition}[theorem]{Definition}
\newtheorem{lemma}[theorem]{Lemma}
\newtheorem{corollary}[theorem]{Corollary}
\newtheorem{remark}[theorem]{Remark}
\crefname{algorithm}{algorithm}{algorithms}
\Crefname{algorithm}{Algorithm}{Algorithms}

\title{Towards Faithful Data Selection:\\ Statistical Limits and Scalable Algorithms}
\author{%
  \begin{tabular}{c@{\hspace{3em}}c}
    \textbf{Ahmed Mehdi Inane}\thanks{Equal contribution First Authors.}
      & \textbf{Mahdi Haghifam}\footnotemark[1]\ \thanks{Part of the work was done while MH was at the Simons Institute for the Theory of Computing, UC Berkeley.} \\
    Université de Montréal, Mila 
      & Toyota Technological Institute at Chicago \\[1.2em]
    \textbf{Gintare Karolina Dziugaite}
      & \textbf{Ioannis Mitliagkas} \\
    Google DeepMind
      & Université de Montréal, Mila, Google DeepMind
  \end{tabular}%
}

\date{}

\usepackage{titlesec}
\titlespacing{\section}{0pt}{\parskip}{0pt}
\titlespacing{\subsection}{0pt}{\parskip}{0pt}
\titlespacing{\subsubsection}{0pt}{\parskip}{0pt}
\usepackage[letterpaper,margin=0.85in]{geometry}
\begin{document}
\maketitle

\begin{abstract}
What does it mean to have an ``optimal'' data selection algorithm? Optimality is conventionally framed as finding a subset that minimizes the population error of the classifier trained only on that subset. Our starting point in this work is first to show that this goal is information-theoretically impossible unless the data selection algorithm has access to a validation set order-wise larger than the training set. This impossibility result motivates shifting from unverifiable population risk minimization to a one-sided guarantee that we refer to as \textit{faithfulness}: for any candidate subset, we provide an estimable proxy that upper-bounds the difference between training on the full training data and on only the subset. We model training with the Unadjusted Langevin Algorithm and bound the Kullback--Leibler divergence between subset-trained and full-data-trained parameter distributions by the trajectory-accumulated gradient discrepancy. Minimizing this discrepancy gives a principled way to select a subset. %
Even though the resulting problem is NP-hard, we propose a swap-based local search that decreases the bound at every accepted step. It can therefore refine the output of any selector without worsening its bound. Finally, we provide experimental results on CIFAR10 and CIFAR100 showing that it matches existing data selection baselines.  Implementation of our algorithm and the code for reproducing the results are available at: \url{https://github.com/Mehdi-Inane/data_selection_swap}.
\end{abstract}

\input{introduction}

\input{related-works-notation}
\input{oracle_vs_trainset}
\input{kl_selection}

\input{experiments}
\input{conclusion}

\bibliography{references,references_2}

\appendix
\input{appendix}

\end{document}

%% file: math_commands.tex
\usepackage{amsmath,amsfonts,bm}

\def\eqref#1{equation~\ref{#1}}

\def\1{\bm{1}}

\DeclareMathAlphabet{\mathsfit}{\encodingdefault}{\sfdefault}{m}{sl}
\SetMathAlphabet{\mathsfit}{bold}{\encodingdefault}{\sfdefault}{bx}{n}

\newcommand{\E}{\mathbb{E}}

\newcommand{\Var}{\mathrm{Var}}

\DeclareMathOperator*{\argmax}{arg\,max}
\DeclareMathOperator*{\argmin}{arg\,min}

\newcommand{\Alg}{\mathcal A}
\newcommand{\Astar}{\Alg^*} %

\newcommand{\F}{\mathbb{F}}

\renewcommand{\Pr}{\mathbb{P}}
\newcommand{\pmask}{p_{\text{masked}}}
\newcommand{\Pfam}{\mathfrak{P}_{\lambda,n,v}}
\newcommand{\ind}{\mathbf 1}
\newcommand{\ncls}{n_{\text{class}}}
\newcommand{\kg}{k_{\text{group}}}
\newcommand{\pres}{p_{\text{resolved}}}

%% file: introduction.tex
\section{Introduction}

Training on ever-larger datasets is expensive, and not all samples contribute equally: careful pruning can improve how performance scales with dataset size \citep{sorscher2023neuralscalinglawsbeating}. \emph{Data selection} formalizes this idea: given a training set $S$ of $n$ samples and a learning algorithm $\Alg$, find a subset $I \subseteq S$ of size $m < n$ such that training on $I$ alone performs as well as training on all of $S$. Besides reducing computation and storage, selection is a concrete instance of the data-centric view of machine learning \citep{zha2025data}, in which the dataset, rather than the model, is the object to curate.

There is a vast body of work on data selection algorithms \citep{moser_coreset_2025}, which differs in what the data selection
algorithm has access to and in how it uses this information.  Bi-level optimization approaches \citep{killamsetty2021glistergeneralizationbaseddata,xia2024less,nikdan2026efficient} select subsets of the training set based on the error on a held-out validation set, and training-data based methods use \emph{only} information stemming from the training set and the learning dynamics induced by a learning algorithm \citep{toneva2019empiricalstudyexampleforgetting,paul_deep_2021,killamsetty_grad-match_2021,cohen2026active}. Despite this empirical progress, the statistical limits of data selection remain poorly understood.

There has also been a recent line of theoretical work on the statistical aspects of data
selection. For instance, data memorization is inherently related to data selection, since memorization
identifies training examples that a learner cannot afford to discard. The
work of \citet{feldman_does_2020} shows theoretically that on long-tailed distributions,
rare examples may provide the only information about subpopulations that collectively
carry substantial test mass, making their memorization necessary for near-optimal
accuracy. Empirically, \citet{feldman_what_2020} show that on standard image benchmarks
memorized examples often improve accuracy on related test points, so dropping rare
samples may degrade the test accuracy. In recent work, \citet{pmlr-v291-hanneke25a} study the
following fundamental question: for empirical risk minimizers, given a set of training
samples, does there exist a (smaller) \emph{good} subset such that training on it performs nearly
as well as training on all the samples? Their main result is to give affirmative answers in various settings. 
Another line of work \citep{kolossov_towards_2023} identifies settings in which selection improves over full-sample training, and provides settings where popular selection methods like unbiased reweighting and influence-based rules are suboptimal. Neither work, however, asks what a selector can identify from the samples alone. %
In this work, we complete this picture by asking: when the training sample contains a subset on which training yields an accurate model, what can a data selection algorithm infer about this subset from finite training (and possibly validation) samples?

To make this question precise, let $S=(Z_1,\dots,Z_n)$ be i.i.d.\ samples from an unknown distribution $P$, and let $\Alg$ be a learning algorithm. For $I\subseteq[n]$, write $R_P(\Alg(S_I))$ for the population risk under $P$ of the
model trained on $S_I=(Z_i)_{i\in I}$, and $R_P^\star$ for the Bayes error. Given a budget $m$, let $\widehat I\subseteq[n]$ with $|\widehat I|= m$ be the subset
chosen by our data selection algorithm, and let
$I^\star\in\arg\min_{I\subseteq[n],\,|I|= m}R_P(\Alg(S_I))$ be the best subset
with the same budget. 
Writing $\Alg(S_{\widehat I})$ to denote the hypothesis learned from $S_{\widehat I}$,
we consider two ways of decomposing the excess risk, $R_P(\Alg(S_{\widehat I})) - R_P^\star$:
\begin{align}
  \underbrace{R_P(\Alg(S_{I^\star})) - R_P^\star}_{\text{compression error}}
   &+ \underbrace{R_P(\Alg(S_{\widehat I})) - R_P(\Alg(S_{I^\star}))}_{\text{selection error}}, 
   \label{eq:decompose-best-subset}
   \\
   \mathllap{\text{ \tiny and }}
  \underbrace{R_P(\Alg(S)) - R_P^\star}_{\text{full-data excess risk}}
   &+ \underbrace{R_P(\Alg(S_{\widehat I})) - R_P(\Alg(S))}_{\text{faithfulness gap}}. 
   \label{eq:training-decomposition}
\end{align}
The first decomposition separates two questions: whether the realized training sample
contains a good subset of size at most $m$ (a question that \citet{pmlr-v291-hanneke25a} studies), and whether the data selection algorithm can identify one.  In our statistical setting, however, the comparator $\Alg(S_{I^\star})$ is not available to the data selection algorithm,
because $I^\star$ depends on the unknown distribution $P$. The second decomposition
replaces this inaccessible comparator with the full-data model $\Alg(S)$; both
$\Alg(S_{\widehat I})$ and $\Alg(S)$ can be trained on the observed data.
This motivates two questions:
\textbf{(1)} When one or more good subsets of size $m$ exist, can a data selection
algorithm find a good subset, and do so using only the training sample and a validation set of moderate size?
\textbf{(2)} When this is impossible, can the algorithm
use the training data alone to ensure faithfulness to
full-data training, and can it optimize this faithfulness
at scale?

\subsection*{CONTRIBUTIONS}
\begin{itemize}[leftmargin=4mm]
  \item \textbf{A statistical limit on the selection error.} We separate the
  existence of a good subset from the ability of a data selector to find it.
  For every compression fraction $\lambda\in(0,1)$, training-set size $n$, and
  validation-set size $v$, we construct a finite family
  $\mathfrak{P}_{\lambda,n,v}$ of zero-Bayes-error classification problems with
  two properties (\cref{thm:one-sample-oracle-separation}). First, the family
  is not degenerate: there is a learner $\Astar$ that, on every problem in the
  family, attains vanishing excess risk both when trained on all $n$ examples
  and when trained on its best subset of size $m=\lfloor\lambda n\rfloor$, so an
  accurate subset always exists. Second, no selector can find one: for every
  learner $\mathcal{A}$ and every selector $\Pi$ that sees only the training
  sample and a validation set of size $v$, some problem in the family forces
  \begin{align*}
     \mathbb{E}\big[\underbrace{R_P(\mathcal{A}(S_{I^\star}))-R_P^\star}_{\text{compression error}}
     +&\underbrace{R_P(\mathcal{A}(S_{\widehat I}))-R_P(\mathcal{A}(S_{I^\star}))}_{\text{selection error}}\big]\\
     &\;\ge\;\Omega_\lambda\!\left(\frac{n}{n+v}\right).
  \end{align*}
  Hence every learner whose compression error vanishes, $\Astar$ among them,
  pays a selection error of order $n/(n+v)$ with every selector: a validation
  set of $v/n\to\infty$ is required to make it vanish.

  \item \textbf{Faithfulness, with a training-data only proxy.} 
  The negative result points to the second decomposition: we ask for a \emph{faithful} subset, one whose training approximates the same distribution over models as training on $S$.  
  By modeling training dynamics of gradient-based methods with the Unadjusted Langevin Algorithm (ULA), we bound the faithfulness by the trajectory-accumulated gradient discrepancy $G_T(I)$ (\cref{thm:path-kl}). 
  This gives a \emph{faithfulness proxy} for any candidate subset. 
  In \cref{fig:corr_d_alpha} we show that the trajectory gradient discrepancy $G_T$ correlates with the faithfulness gap for SGD-trained models.

 \item \textbf{A simple algorithm with an improvement guarantee.}  Minimizing $G_T(I)$ is NP-hard, and we do not claim to solve it optimally.
  Existing gradient-based selectors only optimize relaxations of it, for instance by
  matching gradients at a single training iterate or by scoring examples
  independently (Remark~\ref{rem:umbrella}), so none directly target the quantity our
  bound controls.
  Instead, we give a simple swap-based local search algorithm that, 
  starting from any subset, halts or identifies a subset $I$ that 
  decreases $G_T(I)$. 
  One can start from the full training data, or aim to refine the output of  another selector
  (Theorem~\ref{thm:simple-swap-gap}). Its cost is dominated by per-example gradients, which
  we approximate using a single early checkpoint, yielding an estimate $\widehat{G}_T$ of $G_T$. On CIFAR-10 and CIFAR-100, this simple method is on par with existing static data selection baselines. This opens up avenues for improving the optimization of $G_T(I)$ as a way to improve data selection.

\end{itemize}

%% file: oracle_vs_trainset.tex
\section{Statistical Limits of Optimal Subset Selection}
\label{sec:train_val}

In this section, we start by defining the task of finding the optimal subset in
the training set. A natural comparator (or goal) is as follows: Fix a possibly randomized
learning algorithm $\Alg = \{\Alg_n\}_{n \in \mathbb{N}}$, where $\Alg_m$ is the learner applied to samples of size $m$ and a training sample
$S=(Z_1,\ldots,Z_n)\sim P^n$. Assume that we have a budget of size $m$ for data
selection. For every index set $I\subseteq[n]$, let $S_I=(Z_i)_{i\in I}$, with
the indices in increasing order. Let
\begin{equation} \label{eq:best-subset-pop}
    I_m^\star
    \in
    \argmin_{\substack{I\subseteq[n]\\ |I|= m}}
    R_P\!\left(\Alg_m(S_I)\right),
\end{equation}
with ties broken canonically. Thus $R_P(\Alg_m(S_{I_m^\star}))$ is the risk of the
best subset that an oracle with access to $P$ can find in the realized sample.
Importantly, the oracle must use the same learner $\Alg$ as the data
selector.

Let $V\sim P^v$ be an independent validation set, 
and let $\Pi:\mathcal Z^n\times\mathcal Z^v
    \longrightarrow
    \{I\subseteq[n]:|I|= m\}$ be a (possibly randomized)
data selection algorithm. 
Write $\widehat I_m:=\Pi(S,V)$ for the selected subset. Letting $\Alg = \Alg_m$ denote the learning algorithm fed $m$ samples, 
the excess risk of the hypothesis $\Alg(S_{\widehat I_m})$ can be decomposed into the compression error and selection error as in \cref{eq:decompose-best-subset}.
Both terms in this decomposition are nonnegative: the first asks whether the training sample contains a good
size-$m$ subset for $\Alg_m$, and the second measures the cost of finding
one using $\Pi$. As we will show, there exists a class of problems for which we have a dichotomy: every size-m subset has large compression error for $\Alg_m$, or a
good subset exists, but $\Pi$ cannot identify it.

\begin{theorem}
\label{thm:one-sample-oracle-separation}
Consider multiclass classification with ordinary $0$--$1$ loss.
Fix any compression fraction $\lambda\in(0,\tfrac13)$, set $\gamma:=\tfrac13-\lambda$, and set $K:=2^{\left\lceil 8/\gamma\right\rceil}$. Let the label space be $[K]$. There exist
$n_\lambda\in\mathbb N$ and $C_\lambda>0$, depending only on
$\lambda$, with the following property:
For every integer $n\ge n_\lambda$ and integer validation size $v\ge0$, setting $m=\lfloor\lambda n\rfloor$,
there is a finite
family $\mathfrak P_{\lambda,n,v}$ of deterministic-label
$K$-class classification distributions, together with an explicit
deterministic learner $\Astar$, such that, 
(1) every member of the family has zero Bayes risk and the risk of the model learned by $\Astar$ converges to zero;
(2) for every pair consisting of a learner $\Alg_m$ and a
data selector $\Pi$, there exists $P\in\mathfrak P_{\lambda,n,v}$
such that, for independent samples $S\sim P^n$ and $V\sim P^v$,
\begin{align}
    &\mathbb E_S\!\left[
        R_P\!\left(\Alg_m(S_{I_m^\star})\right)
        -R_P^\star
      \right]\notag\\
    &+
      \mathbb E_{S,V,\Pi}\!\left[
        R_P\!\left(\Alg_m(S_{\widehat I_m})\right)
        -R_P\!\left(\Alg_m(S_{I_m^\star})\right)
      \right]\notag\\
    &\ge
    \frac{\lambda\gamma^2}{264}\,\frac{n}{n+v},
    \label{eq:universal-selection-approximation-lb}
\end{align}
where $\widehat I_m=\Pi(S,V)$ satisfies $|\widehat I_m|= m$,
and $I_m^\star$ minimizes
$R_P(\Alg_m(S_I))$ over $I\subseteq[n]$ with $|I|= m$.
Moreover, these families and the corresponding learners can be
chosen so that, for every $n\ge n_\lambda$ and for every $P \in \mathfrak P_{\lambda,n,v}$
\begin{align}
    \frac{n+v}{n}\,
    \mathbb E_S\!\left[
      R_P\!\left(\Astar(S)\right)
      -R_P^\star
    \right]
    &\le \frac{C_\lambda}{\log n},
    \label{eq:selection-full-data-bound}\\
    \frac{n+v}{n}\,
    \mathbb E_S\!\left[
      R_P\!\left(\Astar(S_{I_m^\star})\right)
      -R_P^\star
    \right]
    &\le \frac{C_\lambda}{\log n}.
    \label{eq:selection-oracle-bound}
\end{align}
In \cref{eq:selection-oracle-bound}, $I_m^\star$ is the optimal
subset for $\Astar$.
\end{theorem}

The full proof is given in \cref{app:construction-proof}. Let us unpack the implications of this result:

     \textbf{Accurate subsets exist}: Bayes risk is zero, and there exists an algorithm for the class of problems ($\Astar$ in \cref{thm:one-sample-oracle-separation}) that achieves vanishing error on both the full sample and a distribution-aware oracle subset of at most $\lfloor\lambda n\rfloor$ examples. Thus the problem is non-degenerate.

    \textbf{Sample-based selection is hard}: Finding such a subset can nevertheless be statistically
    hard: for fixed $\lambda$ and $v=O(n)$, for every selector-algorithm pair such that the algorithm has small compression error, the selection error is bounded below by a constant even with unlimited computation.

    \textbf{Tightness}: Our lower bound in \cref{thm:one-sample-oracle-separation} is tight in its dependence on the number of training samples. Specifically, conditional on the training sample $S$, running $\Alg_m$ on all subsets of $S$ of size $m$ induces a finite hypothesis class of size at most $(en/m)^m$. Standard generalization bounds \citep{shalev2014understanding} guarantee that a validation set of size $\displaystyle v = \Theta(\lambda n/\varepsilon^2)$ suffices to achieve selection error $\varepsilon$.

\begin{remark}[long-tail distributions in practice]
The family of distributions in our lower bound shares the long-tail structure of \citet{feldman_does_2020}: there are as many rare subpopulations as training examples, so many of them appear
once or never in the training set. There, the subpopulations are unrelated, so one that is absent from the training set cannot be predicted, even by full-data
training or by an oracle that knows the distribution. Our main technical contribution is to relate rare subpopulations to one another. This is why an accurate subset of size $m$ exists: an
oracle that knows which rare subpopulations will matter at test time keeps only the examples needed to reconstruct them. 
\end{remark}

\vspace{-1.5em}
\begin{proof}[Proof overview of \cref{thm:one-sample-oracle-separation}]

The natural starting point is to consider a long-tailed distribution from \citet{feldman_does_2020} with $\Theta(n)$ rare subpopulations, so that accurate prediction requires retaining a large amount of information from the training sample. The immediate difficulty is twofold: in a sample of size $n$, a constant fraction of the subpopulations never appear at all, so even full-data training fails on them; and if predicting well requires storing more than $m$ pieces of information, then a population-aware oracle with budget $m$ fails as well. Our construction resolves this difficulty by identifying the subpopulations with the edges of a cubic graph $G$ of logarithmic girth (using Erd\H{o}s--Sachs) and coupling their labels through parity constraints: each edge carries a label in $\mathbb{F}_2^B$, and the labels of the three edges at every vertex sum to zero. 

To create a gap between the oracle and the selector, we partition the edges into small groups and designate one hidden \emph{active} edge in each group; the test distribution places its hard inputs only on active edges. Most examples are \emph{clear}: they reveal a uniformly random edge together with its label. A small fraction are \emph{masked}: their input shows an active edge but hides its label. Since masked examples are rare, most groups receive none of them in $(S,V)$, and for these groups the selector cannot tell which edge is active.

This is what makes selection hard. The parity constraints impose only $2N/3$ equations on $N\approx n$ edges, so the labels still carry about $\frac13 NB$ bits of information. A selector that does not know the active edges must preserve all of it, yet a subset of $m=\lfloor\lambda n\rfloor$ examples conveys at most $n+mB$ bits. When $\lambda<1/3$, this gap forces any selector--learner pair to err on $\Omega(\gamma n)$ edges, and since the active edge of each group is hidden, these errors translate into test risk.

The same graph structure allows full-data training and the oracle to succeed. Each edge is missing from $S$ with probability about $e^{-1}<1/2$, so in the cubic graph $G$ the missing edges form small components, which are trees because $G$ has large girth. Summing the parity constraints over one side of such a tree expresses the label of a missing edge as a sum of observed labels on its boundary. Every edge therefore has a small \emph{certificate} in $S$ that determines its label, except with probability $O(1/\log n)$. Full-data training contains every certificate, while an oracle that knows the active edges keeps only theirs, which fit within the budget $m$.

\end{proof}

%% file: kl_selection.tex
\section{Faithful data selection}\label{sec:kl-selection}

The previous section studied the conventional goal of data selection: returning a
subset of the training sample whose population error nearly matches that of the best
subset $I^\star_m$ in \cref{eq:best-subset-pop}. However, our results show that for
every pair consisting of a learning algorithm and a data selection algorithm, there
exists a scenario in which the data selection algorithm incurs a non-vanishing selection error unless it has
$\omega(n)$ validation samples, even though a simple learner trained on all of $S$
succeeds there. We therefore change the goal.

The selection error in \cref{eq:decompose-best-subset} compares the chosen subset with
$I^\star_m$, which depends on the data distribution $P$ and is unknown to the selector. Instead, let us consider comparisons to the model trained on all of $S$, which the selector can access.
In particular, consider the decomposition of the excess risk of the selected subset described in \cref{eq:training-decomposition}.
The first term (full-data excess risk) does not involve the selector and is typically small in practice, since
data selection is usually applied when training on the full sample already performs
well. Therefore, excess risk of the selected subset is then essentially the second term: the faithfulness gap.
Motivated by this, we seek a subset on which training yields a model that closely tracks the full-data
model.

\begin{remark}
We do not claim that we can overcome the fundamental lower bound in \cref{thm:one-sample-oracle-separation}, since the result provides a lower bound on the excess error directly. Indeed, the hard class of distributions we introduced is such that the faithfulness gap is large. When the gap is small, perhaps we can find and optimize faithful subsets. As we show next, we provide a practical proxy that can be computed from the training set and has the property that, if the proxy is small, we know that the faithfulness gap is small. 
\end{remark}

\subsection{From faithfulness to KL divergence}
How can we control the faithfulness gap? The faithfulness gap in
\cref{eq:training-decomposition} is the difference between the averages of one
function, $\theta\mapsto R_P(\theta)$, under two distributions over trained parameters:
the law $\mathcal{A}_n(S)$ of the model trained on $S$ and the law
$\mathcal{A}_m(S_{\widehat I})$ of the model trained on the subset. We cannot evaluate
this function, since $P$ is unknown, but we do not need to: if the two \emph{laws} are close,
every bounded function of $\theta$ has nearly the same average under both, whatever $P$
is. We therefore call a subset \emph{faithful} if training on it produces nearly the
same distribution over models as training on $S$. This requirement is stronger than a
small faithfulness gap, and usefully so: it controls the gap under every data
distribution at once, and it controls every other bounded evaluation of the trained
model, which we call a \emph{query}. Examples include accuracy on a subgroup or on a
rare subpopulation, accuracy under distribution shift, and the loss or prediction at a
single test point as in data attribution \citep{ye_leave-one-out_2024}. The next result, whose proof (\cref{app:kl-control}) is simply based on the variational representation of total variation distance and Pinsker's inequality, formalizes this intuition:
\begin{proposition}\label{prop:kl_control}
  Let $a,b\in \mathbb{R}$ and $n,m \in \mathbb{N}$ such that $1\leq m \leq n$. Let $q: \mathbb{R}^d \to [a,b]$ be a measurable bounded function. Fix $S \in \mathcal{Z}^n$ as a training set of size $n$ and let $I \subseteq [n]$ be an arbitrary subset of $[n]$ such that $|I|=m$. Then, for every randomized algorithm $\mathcal{A}=\{\mathcal{A}_m\}_{m \in \mathbb{N}}$, we have:
    \begin{align*}
        &\Bigl|\,
            \mathbb{E}_{\theta \sim \mathcal{A}_n(S)}[q(\theta)]
            -
            \mathbb{E}_{\theta' \sim \mathcal{A}_m(S_I)}[q(\theta')]
        \,\Bigr|  \notag \\
        &\;\le\;
        \sqrt{\frac{(b-a)^2}{2}\,D_{\mathrm{KL}}\!\bigl(\mathcal{A}_n(S) \;\|\; \mathcal{A}_m(S_I)\bigr)}.
    \end{align*}
    Note that $\mathcal{A}_n$ and $\mathcal{A}_m$ should both share the same initialization distribution.
\end{proposition}

\section{Practical Data Selection using Langevin Algorithm}

\Cref{prop:kl_control} reduces faithfulness to a single number,
$D_{\mathrm{KL}}\bigl(\mathcal{A}_n(S)\,\|\,\mathcal{A}_m(S_I)\bigr)$. How can we turn this result into a faithful data selection algorithm? To select subsets with it, we need two things: an upper bound on this
divergence that can be evaluated based on the training set, and a
form of that bound that can be minimized over subsets.

In this section, we focus on gradient-based learning algorithms that are often used in practice. To operationalize the KL bound, we approximate gradient-based training by the Unadjusted Langevin Algorithm (ULA). This approximation lets us bound the divergence in closed form and then turn the bound into a practical and scalable algorithm.

More precisely, assume we have a fixed loss function $L_S(\theta) = \frac{1}{|S|}\sum_{x \in S} \ell(\theta,x)  $ and let $d$ be the number of parameters. Fix a number of iterations $T\ge1$, step sizes $h_0,\ldots,h_{T-1}>0$, inverse
temperatures $\beta_0,\ldots,\beta_{T-1}>0$. On a training
set $S'$, ULA starts from an initialization distribution $\theta_0 \sim \mu_0$ and, for $t=0,\ldots,T-1$, let
\begin{equation}\label{eq:ula}
  \theta_{t+1} = \theta_t - h_t\nabla L_{S'}(\theta_t)
  + \sqrt{2h_t/\beta_t}\,\xi_{t+1},
\end{equation}
where $\xi_1,\xi_2,\ldots$ are i.i.d.\ $\mathcal{N}(0,I_d)$ and independent of
$\theta_0$. We write $\mu_t^{S'}$ for the law of $\theta_t$, and we take the learner to
be ULA stopped at iteration $T$, that is, $\mathcal{A}_k(S'):=\mu_T^{S'}$ for every
sample $S'$ of size $k$. ULA is gradient descent with injected Gaussian noise, and its
stochastic-gradient version is the SGLD algorithm of \citet{wellingteh}. ULA is an approximation of how models are trained in practice. Our guarantees are proven for ULA, and \cref{sec:experiments} shows that the
resulting selection rule remains effective when models are trained with SGD.

Before stating our results, we define a quantity
that we refer to as the \emph{trajectory gradient discrepancy},
\begin{equation}\label{eq:grad-discrepancy}
  G_T(I) := \sum_{t=0}^{T-1} \frac{\beta_t h_t}{4} \mathbb{E}_{\theta_t \sim\mu_t^{S}}
  \bigl\|\nabla L_{S_I}(\theta_t)-\nabla L_S(\theta_t)\bigr\|^2,
\end{equation}
the squared distance between the subset and full-data gradients, accumulated along the
full-data training trajectory. The proof is deferred to \cref{app:path-kl}.
\begin{proposition}\label{thm:path-kl}
Fix $T\ge1$, step sizes $h_0,\ldots,h_{T-1}>0$, inverse temperatures
$\beta_0,\ldots,\beta_{T-1}>0$, integers $1\le m\le n$, a
training set $S\in\mathcal{Z}^n$, and $I\subseteq[n]$ with $|I|=m$, and assume that
$\ell(\cdot,z)$ is continuously differentiable for every $z\in\mathcal{Z}$. Assume furthermore that $\mathbb{E}_{\theta \sim \mu_t^S}\left[ \|\nabla \ell (\theta,Z_i)\|^2\right] < \infty$ for all $i \in [n], 0<t \leq T-1$. Run ULA on
$S$ and on $S_I$ with these schedules. Then, $D_{\mathrm{KL}}\bigl(\mathcal{A}_n(S)\,\|\,\mathcal{A}_m(S_I)\bigr) \le G_T(I)$.
\end{proposition}

\begin{corollary}[Faithfulness proxy under ULA]\label{cor:ula-faithfulness}
Under the assumptions of \cref{thm:path-kl}, for all $a<b$ and every measurable
$q:\mathbb{R}^d\to[a,b]$,
\[
  \Bigl|\,\mathbb{E}_{\theta\sim\mathcal{A}_n(S)}[q(\theta)]
  -\mathbb{E}_{\theta\sim\mathcal{A}_m(S_I)}[q(\theta)]\,\Bigr|
  \le (b-a)\sqrt{G_T(I)/2}.
\]
\end{corollary}

The path-space identity behind \cref{thm:path-kl} is the same quantity that appears as ``gradient incoherence'' in information-theoretic generalization bounds for SGLD \citep{negrea_information-theoretic_2019,haghifam_sharpened_2020}, and KL divergences between Langevin chains on neighbouring datasets drive Langevin unlearning \citep{chien_langevin_2024,inane2026unlearningasymmetricsourcesimproved} and leave-one-out distinguishability \citep{ye_leave-one-out_2024}. 

\begin{figure*}[htbp]
  \centering
  \begin{subfigure}[b]{0.48\textwidth}
    \centering
    \includegraphics[width=0.8\linewidth]{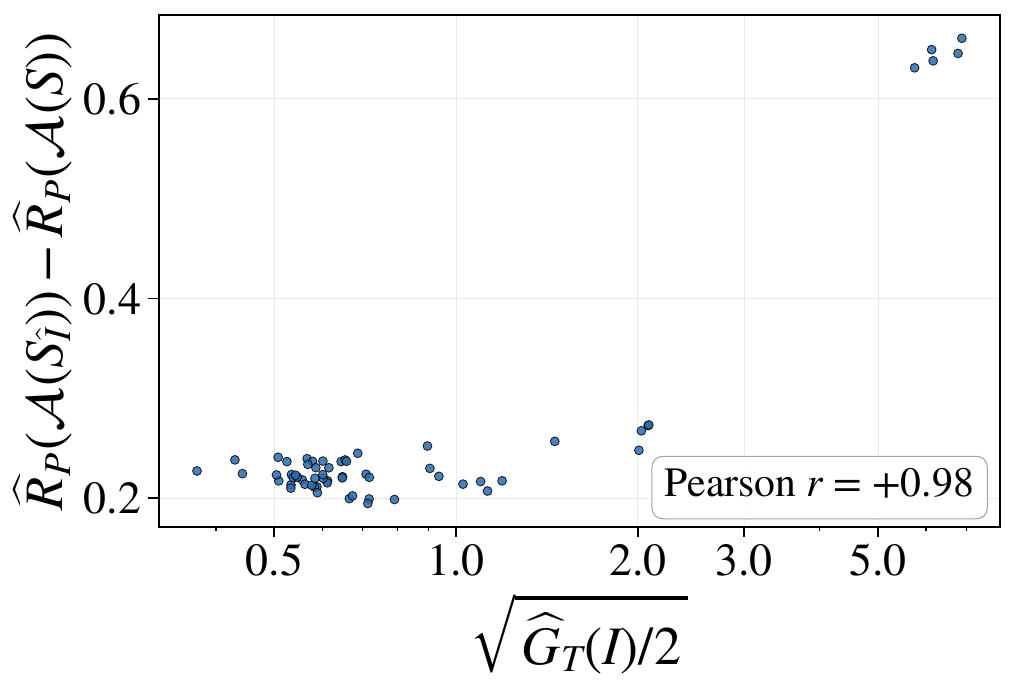}
    \caption{ResNet-18 on CIFAR-10}
    \label{fig:col_left}
  \end{subfigure}
  \hfill
  \begin{subfigure}[b]{0.48\textwidth}
    \centering
    \includegraphics[width=0.8\linewidth]{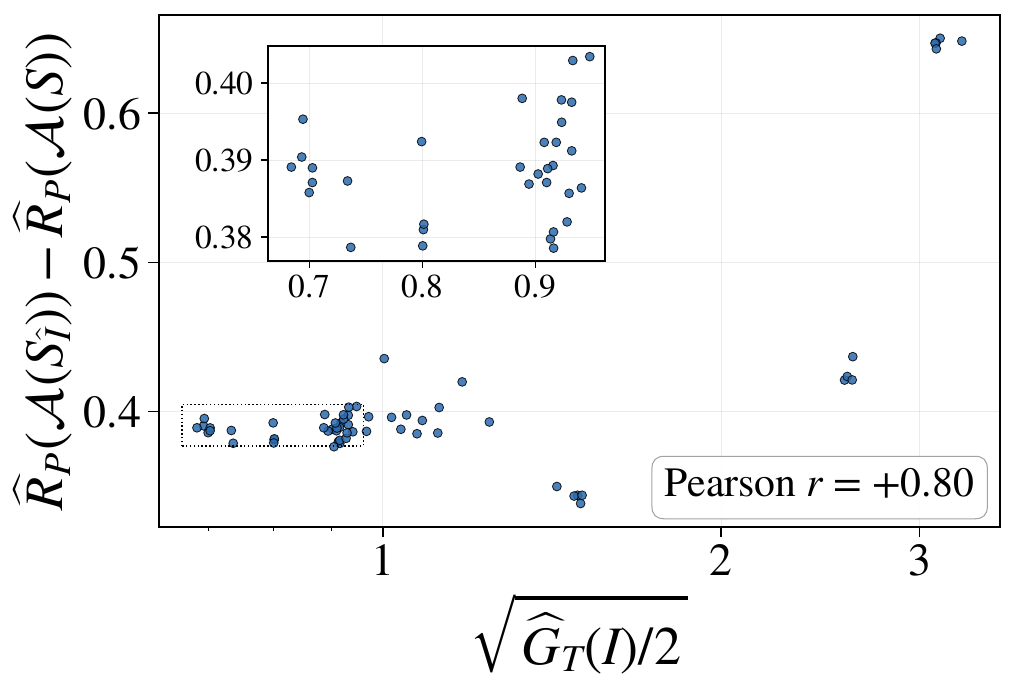}
    \caption{VGG-16 on CIFAR-100}
    \label{fig:col_right}
  \end{subfigure}

\caption{Empirical Faithfulness gap $\widehat{R}_P(\mathcal{A}(S_I)) - \widehat{R}_P(\mathcal{A}(S))$ versus the estimate of $\sqrt{\widehat{G}_T(I)/2}$ for CIFAR-10 / ResNet-18 (left) and CIFAR-100 / VGG-16 (right) at retention fraction $0.1$. 
Each point represents a subset from a data selection method, and $\mathcal{A}$ represents the SGD training procedure. Subsets with large $G_T(I)$ are the ones with the largest gap. The inset (right) zooms in on the dotted region. Points in the upper-right corner correspond to EL2N (EL2N in \cref{fig:baseline_comparison} achieves lower accuracy), confirming the theoretical prediction of  \cref{cor:ula-faithfulness}. 
} \label{fig:corr_d_alpha}
\end{figure*}

Our contribution is not the bound but its use: we read $G_T$ as an estimable, distribution-free \emph{proxy} for the selection objective (we show in \cref{sec:experiments} that an estimate of $G_T$ can predict the downstream test accuracy). A notable property is that the expectation in \eqref{eq:grad-discrepancy} is over the full-data chain only, so $G_T(I)$ can be
estimated for every candidate subset from the same full-data training runs, using per-example gradients; this requires no training on any subset and no access to $P$ or to held-out data. In \cref{fig:corr_d_alpha}, we estimate $G_T$ using a single full training run for a stochastic gradient descent (SGD) trajectory and show that it still correlates with the faithfulness error.

\begin{remark}[$G_T$ and existing gradient-based selectors]
\label{rem:umbrella}
At a single iterate, with $g_i := \nabla\ell(\theta,Z_i)$ and
$\bar g := \nabla L_S(\theta)$,
\[
  \Bigl\|\tfrac{1}{m}\sum_{i\in I} g_i-\bar g\Bigr\|^2
  \;=\; \tfrac{1}{m^2}\sum_{i,j\in I}\langle g_i,g_j\rangle
  \;-\; \tfrac{2}{m}\sum_{i\in I}\langle g_i,\bar g\rangle
  \;+\; \|\bar g\|^2 .
\]
Several selectors optimize pieces of this expansion.
CRAIG~\citep{mirzasoleiman_coresets_2020} and
GRAD-MATCH~\citep{killamsetty_grad-match_2021} minimize the same discrepancy at
the current iterate but over weighted subsets, via a facility-location
surrogate or orthogonal matching pursuit, and GRAD-MATCH re-solves it during
training; $G_T$ is the unweighted, trajectory-averaged version, and
\cref{thm:path-kl} is what connects it to the trained model.
Ranking examples by alignment $\langle g_i,\bar g\rangle$ (TracIn-style scores
with the training loss as target;
\citealp{pruthi_estimating_2020}) optimizes the linear term alone and ignores
the Gram term, i.e.\ redundancy among the selected examples.
GraNd~\citep{paul_deep_2021} ranks by $\|g_i\|$, which enters $G_T$ only through
the diagonal of the Gram term; since GraNd and EL2N keep the largest scores,
they increase this term, which is consistent with their subsets being the least
faithful ones in \Cref{fig:corr_d_alpha}.
\end{remark}

Based on the trajectory gradient discrepancy, we define the following quadratic integer programming problem (more details in \cref{app:qp_formulation}):

\begin{definition}[Selection as a quadratic program]\label{thm:qp_formulation}
In the setting of \cref{thm:path-kl}, write
$\zeta_i(\theta):=\nabla\ell(\theta,Z_i)-\nabla L_S(\theta)$ for the centered
per-example gradients and define $C\in\mathbb{R}^{n\times n}$ by
$
  C_{ij}:=\sum_{t=0}^{T-1}\frac{\beta_t h_t}{4}\,\mathbb{E}_{\theta\sim\mu_t^S}
  \bigl\langle\zeta_i(\theta),\zeta_j(\theta)\bigr\rangle .
$
Then $C$ is positive semidefinite, $C\mathbf{1}=0$, and
$G_T(I)=\frac{1}{m^2}\mathbf{1}_I^\top C\,\mathbf{1}_I$ for every $I\subseteq[n]$ with
$|I|=m$. Consequently, minimizing $G_T(I)$ over subsets of size $m$ is the integer
quadratic program
\[
  \min_{\alpha\in\mathcal{B}_m}D(\alpha)\triangleq\frac{1}{m^2}\alpha^\top C\alpha,
\]
where $\mathcal{B}_m:=\bigl\{\alpha\in\{0,1\}^n:\mathbf{1}^\top\alpha=m\bigr\}$.
\end{definition}

\subsection{Subset selection via gradient swaps}\label{sec:swap}

How can we solve the program in \cref{thm:qp_formulation} when $n$ is large? Three
facts rule out the obvious approaches. 
First, exact minimization of the program in \cref{thm:qp_formulation} is NP-hard even for $T=1$ and $d-1$ by reduction from fixed-cardinality subset sum (and general MIQP is NP complete; \citep{delpia2014mixedintegerquadraticprogrammingnp}).
Second, forming $C$ requires
$\Theta(n^2)$ inner products at every point of the trajectory, and storing it requires
$\Theta(n^2)$ memory. Third, relaxing $\alpha$ to $[0,1]^n$ is uninformative: since
$C\mathbf{1}=0$, the fractional point $\frac{m}{n}\mathbf{1}$ has objective value zero
and says nothing about which subset to select. These observations prompt us to focus on designing an algorithm that requires only
matrix--vector products with $C$, keeps every iterate a subset of size $m$, and never
increases $G_T$. \Cref{alg:gradient-ranked-swap} is such a method: a local search that
exchanges one selected example for one unselected example at a time.

\begin{algorithm}[t]
\caption{Gradient-ranked swap descent}
\label{alg:gradient-ranked-swap}
\begin{algorithmic}[1]
\Require products $v\mapsto Cv$; budget $1\le m\le n-1$; initial subset
  $I_0\subseteq[n]$ with $|I_0|=m$
\Ensure subset $I_{\mathrm{out}}\subseteq[n]$ with $|I_{\mathrm{out}}|=m$
\State $s_0\gets\frac{2}{m^2}\,C\mathbf{1}_{I_0}$
  \Comment{$s_0=\nabla D(\mathbf{1}_{I_0})$}
\For{$k=0,1,2,\ldots$}
  \State $i_k\in\argmax_{i\in I_k} s_{k}(i)$ \Comment{drop}
  \State $j_k\in\argmin_{j\notin I_k} s_{k}(j)$ \Comment{add}
  \State $I'\gets(I_k\setminus\{i_k\})\cup\{j_k\}$ 
   \State $s'\gets\frac{2}{m^2}\,C\mathbf{1}_{I'}$
  \If{$G_T(I')<G_T(I_k)$} \Comment{$G_T(I)=\frac12\mathbf{1}_I^\top s$}
    \State $(I_{k+1},s_{k+1})\gets(I',s')$
  \Else
    \State \Return $I_{\mathrm{out}}:=I_k$
  \EndIf
\EndFor
\end{algorithmic}
\end{algorithm}

\paragraph{Overview of \Cref{alg:gradient-ranked-swap}.} Fix a subset $I$ of size $m$ and let
$s:=\frac{2}{m^2}C\mathbf{1}_I$. Since $\frac1m\sum_{j\in I}\zeta_j
=\nabla L_{S_I}-\nabla L_S$, the entry $s_i$ equals $\frac{2}{m}$ times the trajectory
average, weighted as in \cref{eq:grad-discrepancy}, of
$\langle\zeta_i,\nabla L_{S_I}-\nabla L_S\rangle$: it measures how strongly example $i$
is aligned with the current gradient error of the subset. Exchanging $i\in I$ for
$j\notin I$, so that $I'=(I\setminus\{i\})\cup\{j\}$, changes the objective by exactly
\begin{equation}\label{eq:swap-change}
  G_T(I')-G_T(I)=s_j-s_i+\frac{C_{ii}+C_{jj}-2C_{ij}}{m^2}.
\end{equation}
The pair $(i_k,j_k)$ minimizes the first-order term $s_j-s_i$ over all $m(n-m)$
exchanges: it drops the selected example most aligned with the error and adds the
unselected example most opposed to it. The second term is the weighted squared distance
between the gradients of examples $i$ and $j$ along the trajectory. It is nonnegative,
so the first-order prediction is optimistic, and the algorithm therefore evaluates the
exact objective before accepting the exchange. Every accepted exchange lowers $G_T$, and
with it the faithfulness proxy $\sqrt{G_T(I)/2}$ of \cref{cor:ula-faithfulness}. Whether the
algorithm starts from a uniformly random subset or from the output of another selection
method, it returns a subset whose $G_T$ is at least as small as that of its
starting point. Note that although $G_T$ is defined along a full-data training run, \cref{sec:experiments} shows that a single early checkpoint suffices in practice. 

\begin{remark}\label{rem:cost}
Evaluating $v\mapsto Cv$  requires the per-example gradients of all $n$ pool examples at every checkpoint, or an $n\times n$ matrix built at $\Theta(n^2d)$ cost per checkpoint. Our implementation removes the memory requirement by storing only aggregate gradients.
\end{remark}

\begin{theorem}[Swap descent decreases the faithfulness bound]\label{thm:simple-swap-gap}
In the setting of \cref{thm:qp_formulation}, let $1\le m\le n-1$. For every initial
subset $I_0\subseteq[n]$ with $|I_0|=m$ and every tie-breaking rule in the choice of
$(i_k,j_k)$, \Cref{alg:gradient-ranked-swap} generates subsets $I_0,I_1,\ldots$ of size
$m$ and returns $I_{\mathrm{out}}$ such that:
\begin{enumerate}
  \item[(i)] every accepted swap strictly decreases the objective,
  $G_T(I_{k+1})<G_T(I_k)$, and the algorithm terminates after finitely many
  iterations;
  \item[(ii)] if $I_0$ is drawn uniformly from the subsets of $[n]$ of size $m$, then
  $
    \mathbb{E}\bigl[G_T(I_{\mathrm{out}})\bigr]
    \le \frac{n-m}{m(n-1)}\sum_{t=0}^{T-1}\frac{\beta_t h_t}{4}\,V_t,
  $
  where
  $V_t:=\frac1n\sum_{i=1}^n\mathbb{E}_{\theta\sim\mu_t^S}
  \|\nabla\ell(\theta,Z_i)-\nabla L_S(\theta)\|^2$
  and the expectation is over the initialization.
\end{enumerate}
\end{theorem}
The proof alongside the construction of the algorithm is detailed in \cref{app:simple-swap-gap}.
\paragraph{Relation to prior work.}
$G_T$ is not submodular, so greedy combinatorial algorithms and facility-location surrogates \citep{pmlr-v37-wei15,submodinformation} carry no guarantee for it. Swap descent is thus a way to propose swaps from first-order scores but accepts them using the exact quadratic, so accepted swaps never increase $G_T$ (Theorem~\ref{thm:simple-swap-gap}).

%% file: experiments.tex
\section{Experiments}
\label{sec:experiments}

\begin{figure*}[t]
  \centering
  \begin{subfigure}[b]{0.32\textwidth}
    \centering
    \includegraphics[width=1\linewidth]{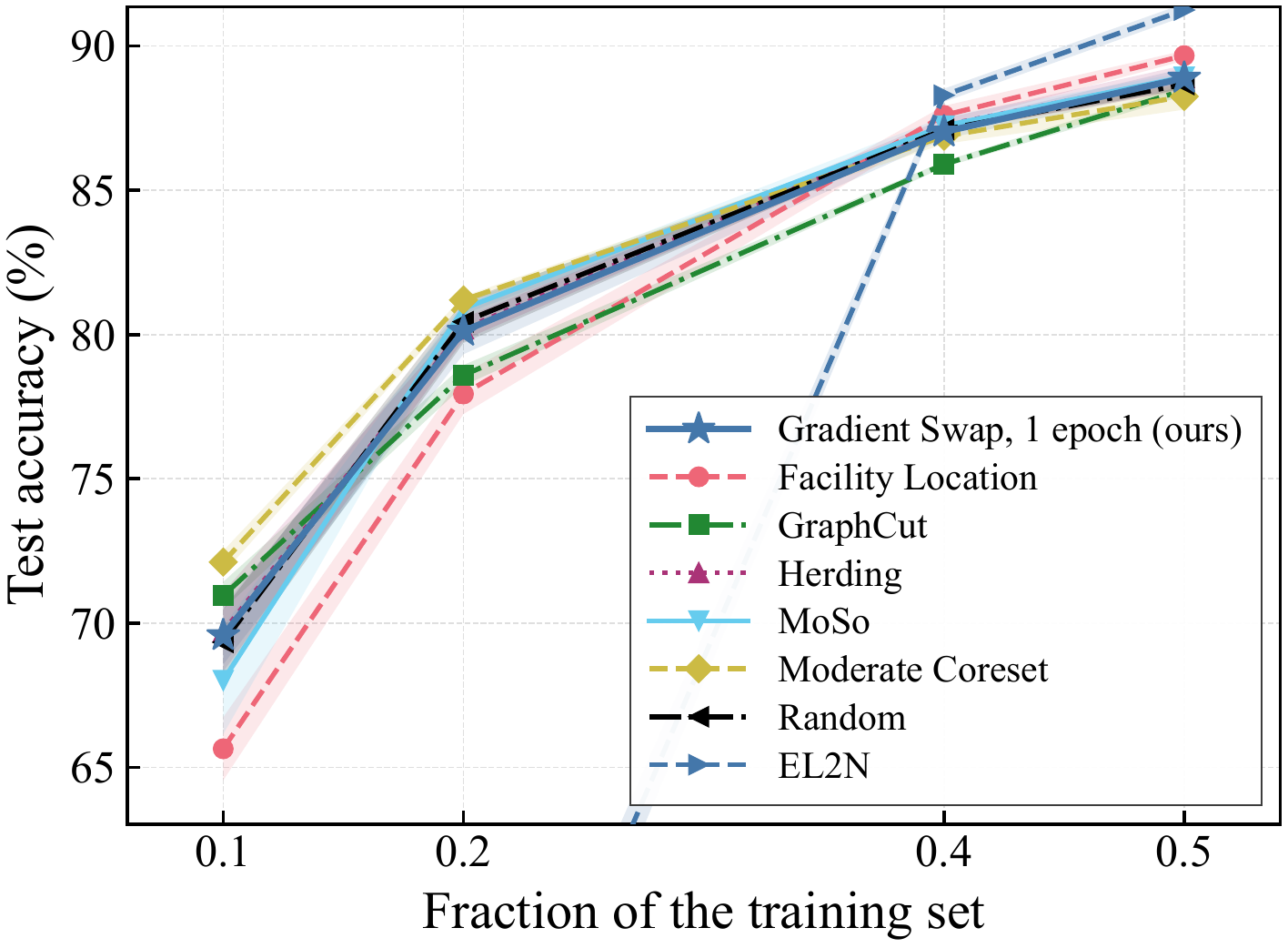}
    \caption{ResNet-18}
    \label{fig:resnet-cifar10}
  \end{subfigure}
  \hfill
  \begin{subfigure}[b]{0.32\textwidth}
    \centering
    \includegraphics[width=1\linewidth]{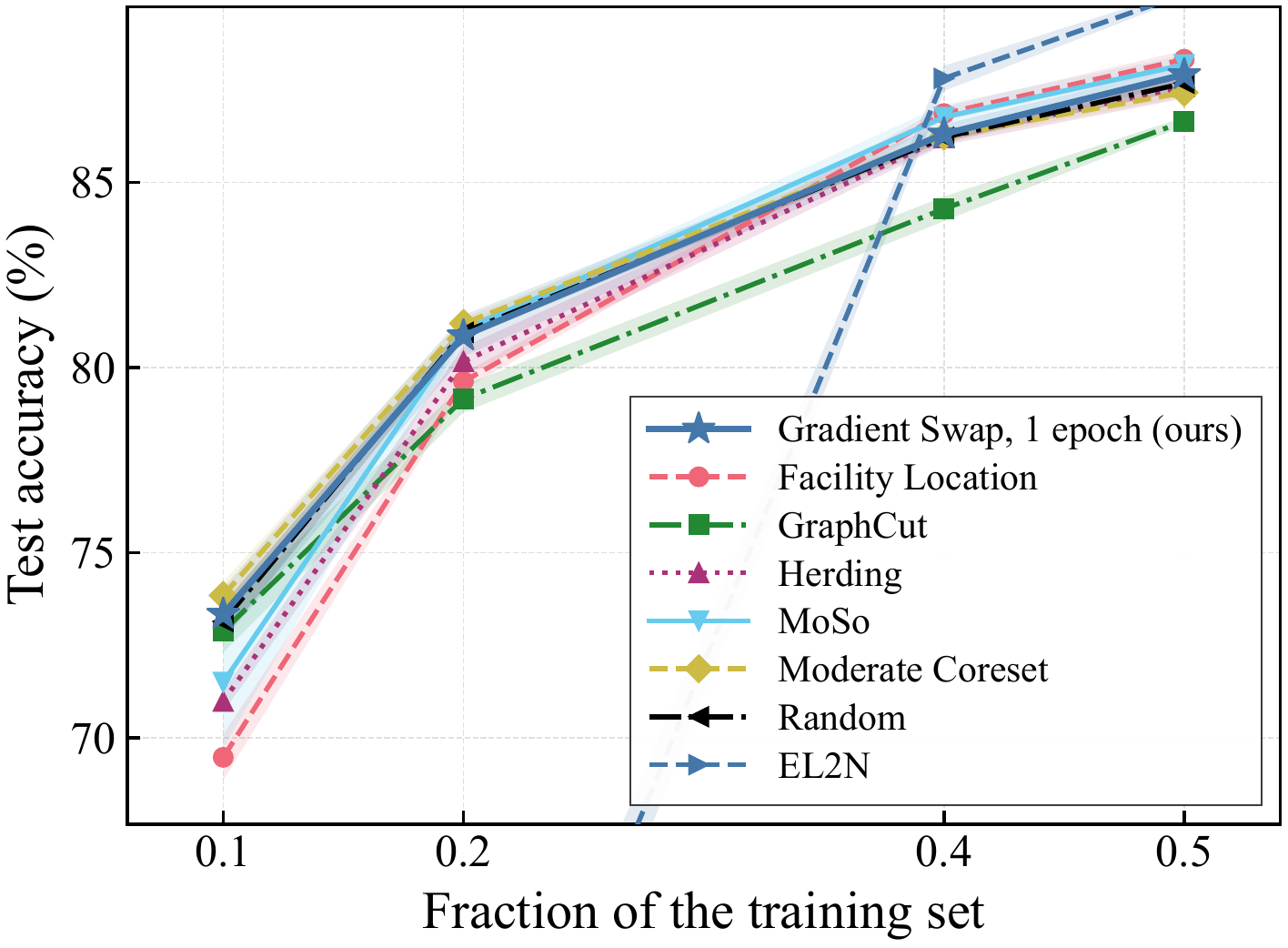}
    \caption{VGG-16}
    \label{fig:vgg-cifar10}
  \end{subfigure}
  \hfill
  \begin{subfigure}[b]{0.32\textwidth}
    \centering
    \includegraphics[width=1\linewidth]{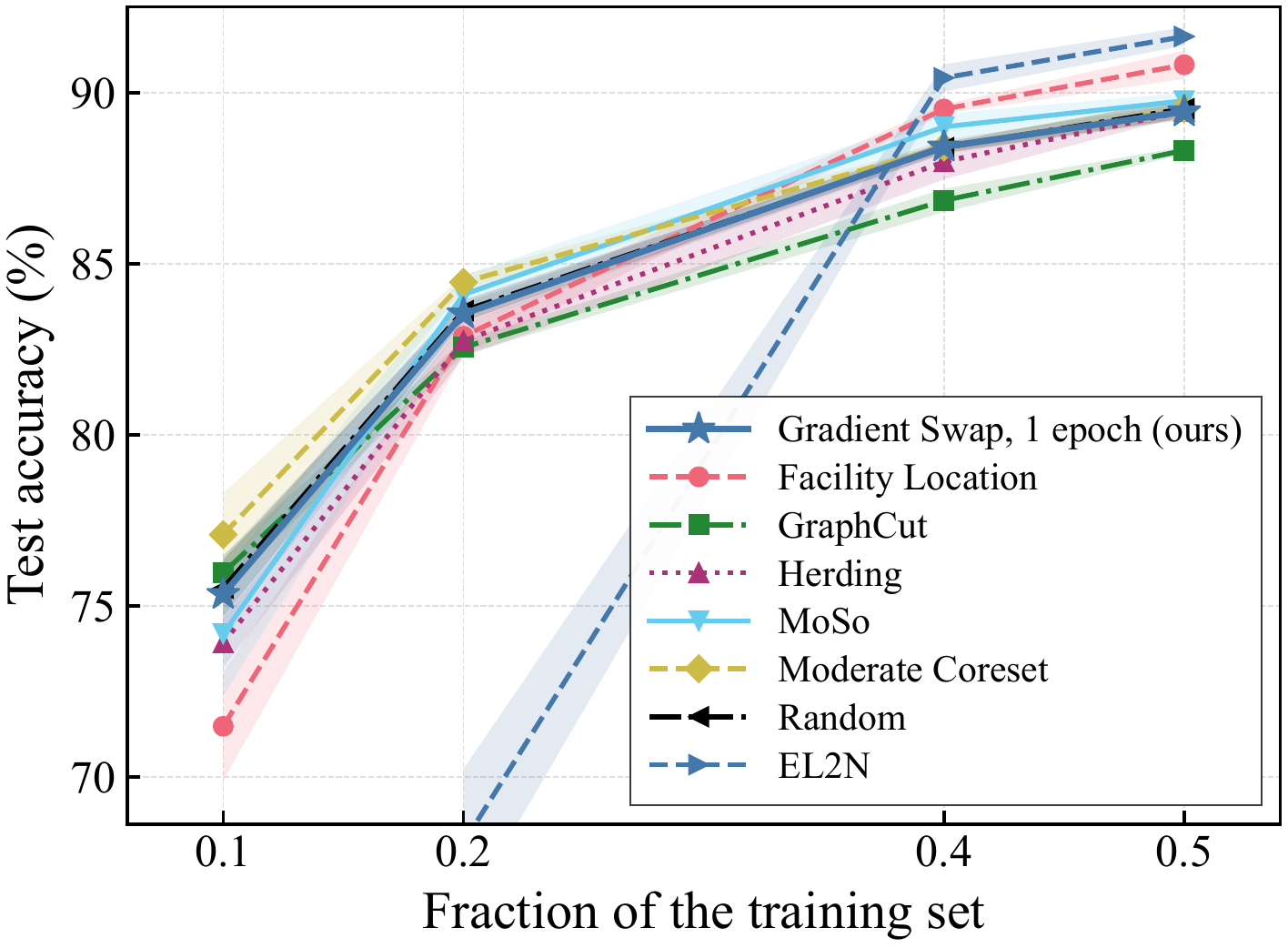}
    \caption{Inception}
    \label{fig:inception-cifar10}
  \end{subfigure}

  \vspace{2pt}
  \begin{subfigure}[b]{0.32\textwidth}
    \centering
    \includegraphics[width=1\linewidth]{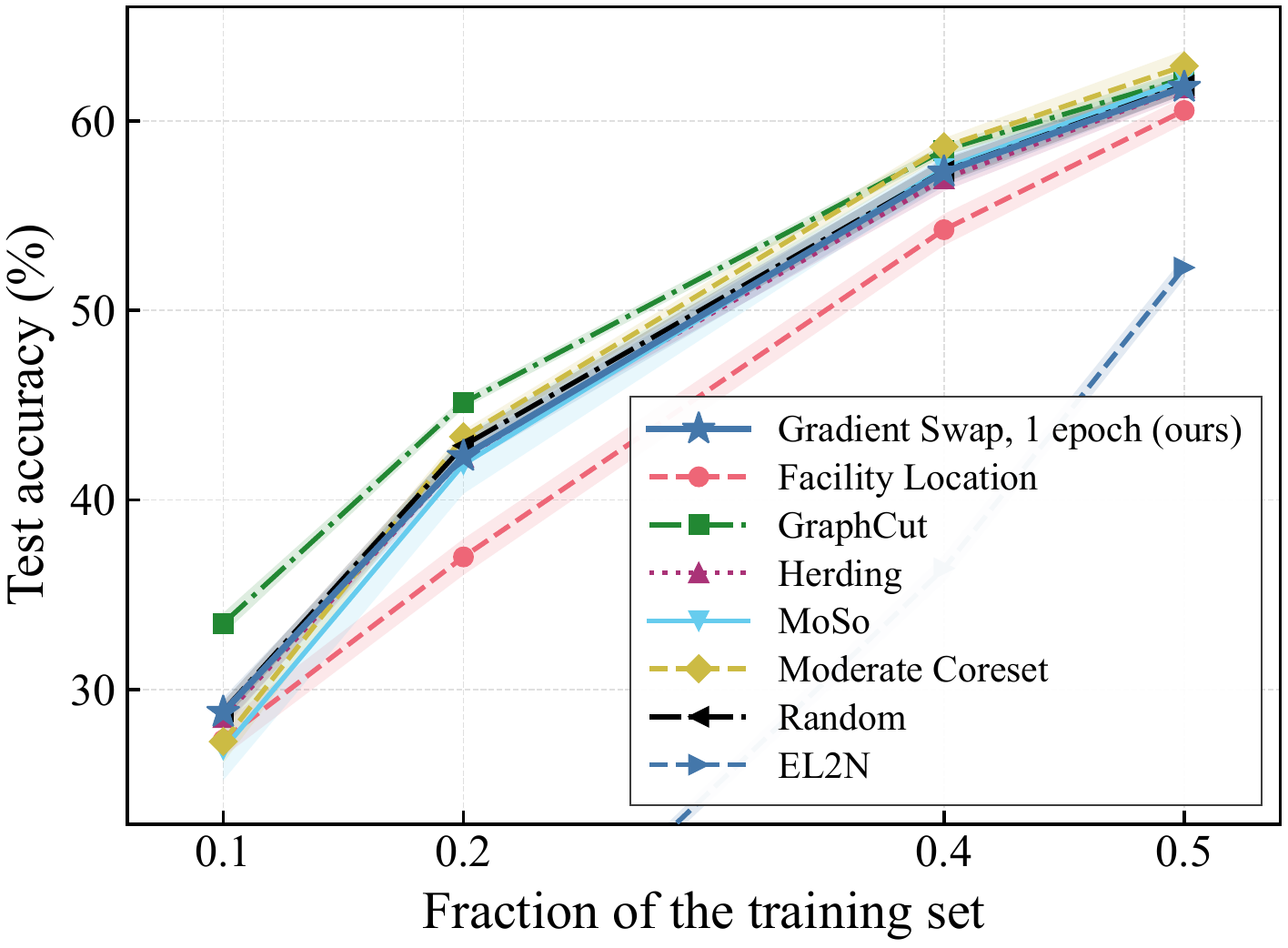}
    \caption{ResNet-18}
    \label{fig:resnet-cifar100}
  \end{subfigure}
  \hfill
  \begin{subfigure}[b]{0.32\textwidth}
    \centering
    \includegraphics[width=1\linewidth]{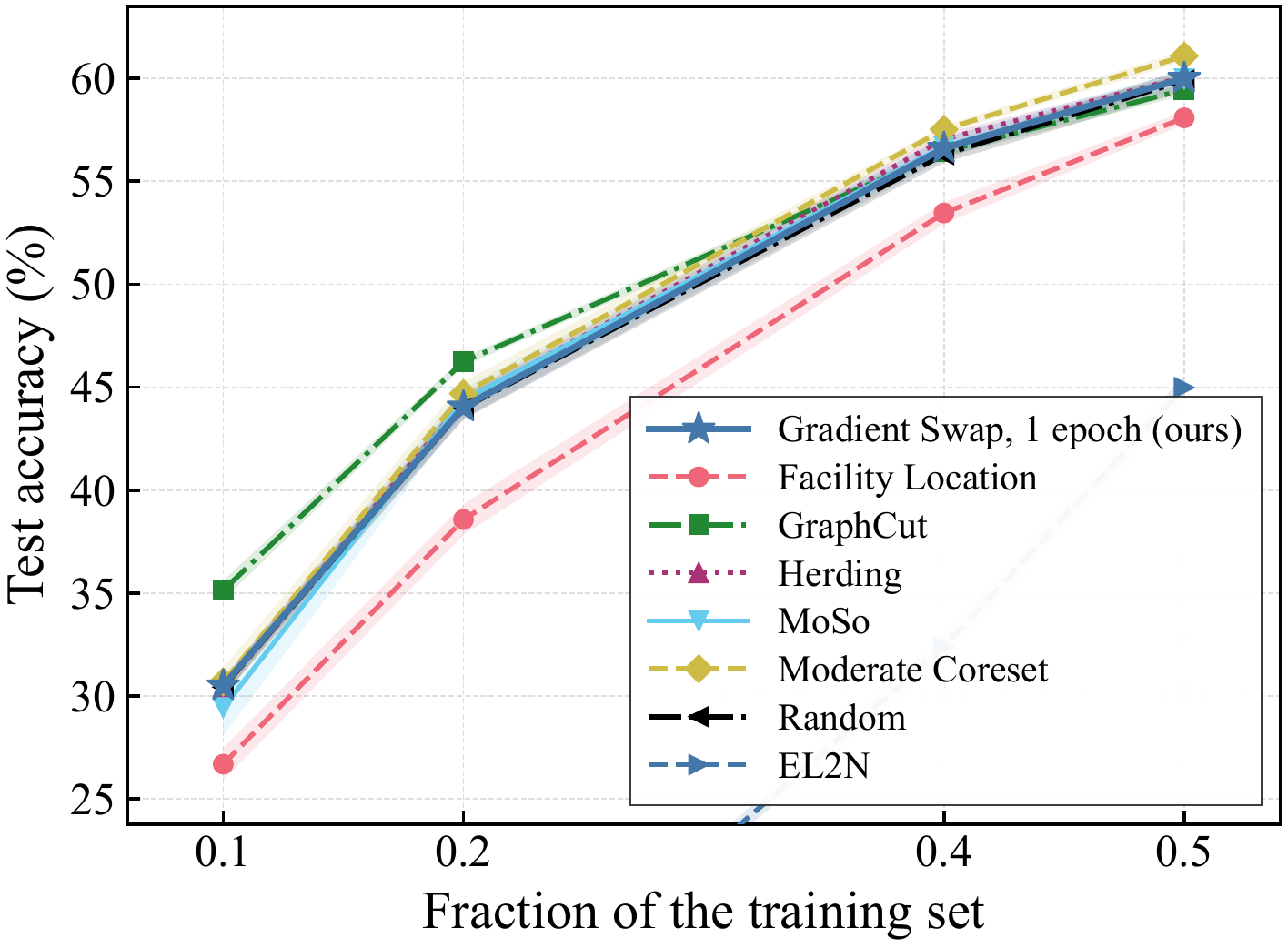}
    \caption{VGG-16}
    \label{fig:vgg-cifar100}
  \end{subfigure}
  \hfill
  \begin{subfigure}[b]{0.32\textwidth}
    \centering
    \includegraphics[width=1\linewidth]{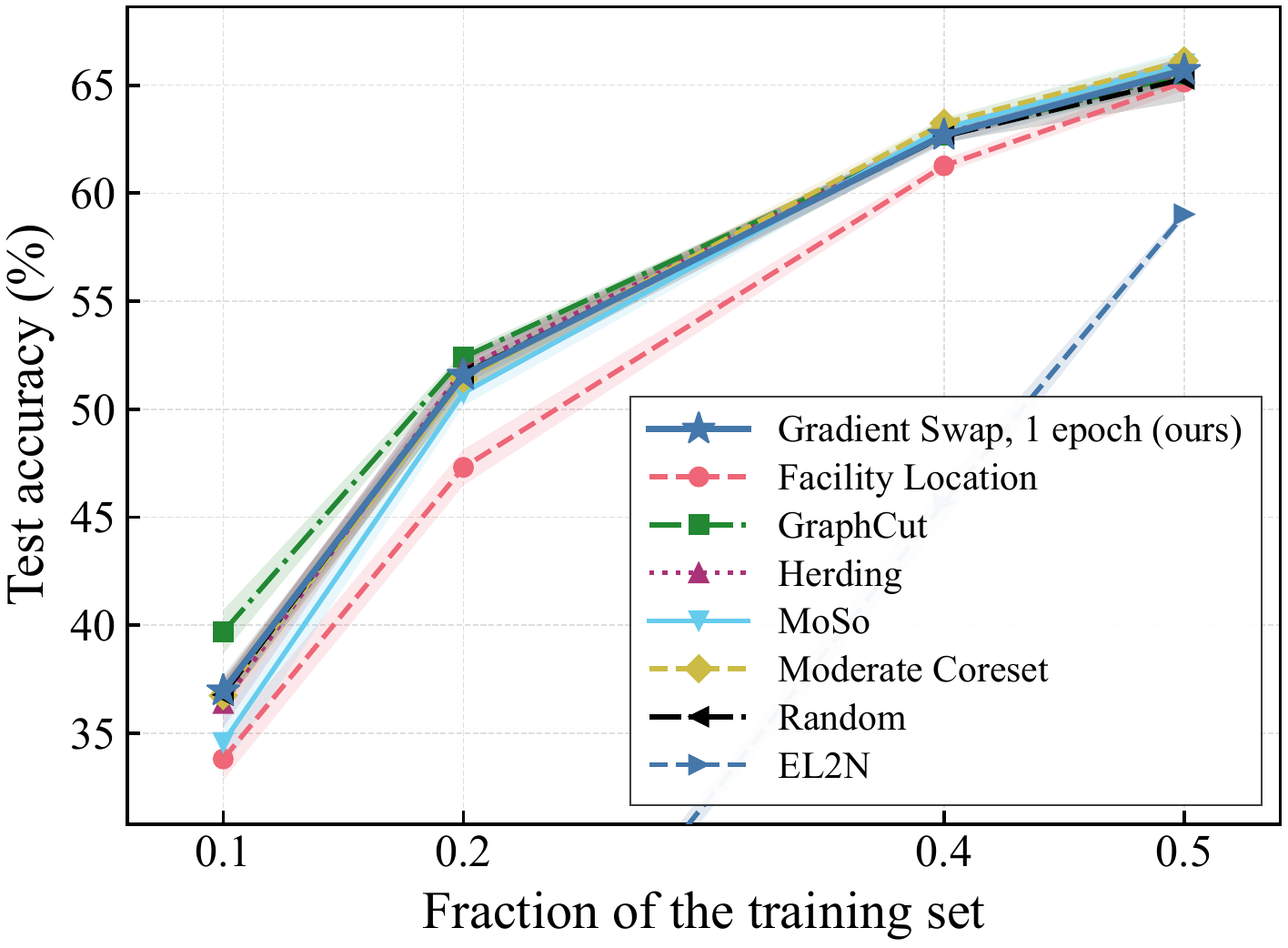}
    \caption{Inception}
    \label{fig:inception-cifar100}
  \end{subfigure}

  \caption{Test accuracy of the retrained model versus retention fraction on
  CIFAR-10 (top row) and CIFAR-100 (bottom row), averaged over five seeds (std in shaded area). Gradient
  Swap uses $\tau=1$ (epoch~1). It is competitive with static baselines while being the
  only method that optimizes the estimated $G_T$.}
  \label{fig:baseline_comparison}
\end{figure*}

We evaluate an SGD-based estimate $\widehat{G}_T$ of $G_T$ and our swap algorithm (Algorithm~\ref{alg:gradient-ranked-swap}) around three questions: (i)~how much of the full-data trajectory is needed to estimate $G_T$? (ii)~does $\widehat{G}_T$ track the quality of the subsets it scores? and (iii)~does optimizing $\widehat{G}_T$ via local search match established selectors in accuracy?

\subsection{Experimental setup}
\label{sec:setup}
\textbf{Datasets and models.}
We use CIFAR-10 and CIFAR-100 \citep{alex2009learning} with three architectures:
ResNet-18 \citep{he2016deep}, VGG-16 \citep{simonyan2015deepconvolutionalnetworkslargescale},
and Inception \citep{szegedy2014goingdeeperconvolutions}. The same datasets and architectures are used for every
selection method. Models are trained using SGD with Nesterov momentum $0.9$, initial learning rate $0.01$, and weight decay $5\times 10^{-4}$. The learning rate follows a cosine schedule for $300$ epochs, with batch size $128$ and cross-entropy loss. The same optimizer and schedule are used both to train the full-data reference trajectory and to retrain on a selected subset.

\textbf{Estimating $G_T$.}
We do not compute $G_T$ exactly; we estimate it and denote the estimate by
$\widehat{G}_T$. We train the full-data reference model once on a fixed random
$90\%/10\%$ train/validation split and store checkpoints $\theta_k$ (caveat in \cref{app:caveat_train_val}). The expectation
over $\mu_t^S$ in the definition of $G_T$ is replaced by this single realized
trajectory, and the sum over steps by a weighted sum over the stored checkpoints. Neither $C$ nor per-example gradients are ever formed. Writing
$u_I^k=\sum_{i\in I}\nabla\ell(\theta_k,Z_i)$ and $\bar g_k=\nabla L_S(\theta_k)$, we
have $\widehat{G}_T(I)=\sum_k w_k\,\|u_I^k/m-\bar g_k\|^2$, so only one
\emph{aggregate} gradient per checkpoint is stored. The cost is still linear in the number of
checkpoints, which makes swaps expensive for long horizons $T$, since aggregate gradients need to be recomputed after each swap. For selection we
therefore use $\tau\in\{1,4,10,15,20\}$ checkpoints: $\tau=1$ is a single-checkpoint
estimate (epoch~1), and larger $\tau$ approaches the full trajectory average of the
theory. In Figure~\ref{fig:corr_d_alpha}, $\widehat{G}_T$ is evaluated
once per subset with no swap sweeps, so we use \emph{all} $300$ stored checkpoints.
We refer to Algorithm~\ref{alg:gradient-ranked-swap} as Gradient Swap and add a
second stopping criterion, a maximum of $200$ allowed swaps.

\textbf{Baselines.}
We compare against six static data-selection methods: Herding \citep{welling2009herding},
Moderate Coreset \citep{xia2022moderate}, Facility Location and GraphCut
\citep{submodinformation}, EL2N \citep{paul_deep_2021}, and MoSo \citep{tan2023data}.
We also include random selection, which draws a uniform subset. We omit Grad-Match \citep{killamsetty_grad-match_2021} and CRAIG \citep{mirzasoleiman_coresets_2020}, since they are \emph{adaptive} methods that reselect the subset periodically during training, whereas we focus on static selection, where a single subset is chosen before training.

\textbf{Evaluation protocol.}
Each method selects a subset at four retention fractions of the training pool
($10\%$, $20\%$, $40\%$, $50\%$), and the model is then retrained on the selected
subset from scratch. Our main metric is the test accuracy of the retrained model,
averaged over five seeds. To assess whether $G_T$ is a meaningful target, we compare the estimated bound $\sqrt{\hat{G}_T(I)/2}$ to the faithfulness gap $R_P(\mathcal{A}(S_I)) - R_P(\mathcal{A}(S))$ for subsets produced by different selection methods at retention $0.1$, where $\mathcal{A}$ is SGD, and report Pearson correlation.

\subsection{Results}
\label{sec:results}

\textbf{A single early checkpoint is enough to estimate $G_T$.}
\Cref{tab:checkpointing} reports the test accuracy of the selected subsets for
$\tau \in \{1, 4, 10, 15, 20\}$. No two settings in a column differ by more than
roughly one standard error of their difference, and there is no monotone trend in
$\tau$. Additional checkpoints thus buy no detectable improvement while costing
proportionally more, so we use $\tau=1$ in all remaining experiments. This echoes the
finding of \citet{paul_deep_2021} that relevant examples can be identified in the
early stages of training. Results for other settings are reported in
\cref{app:checkpoint_ablation}.

\textbf{$\widehat{G}_T$ separates faithful subsets from unfaithful ones.}
\Cref{fig:corr_d_alpha} plots the faithfulness gap against the estimated bound
$\sqrt{\widehat{G}_T(I)/2}$. Subsets with large $\widehat{G}_T$ are the ones with the largest gap, with
a Pearson correlation of $r=+0.98$ for ResNet-18 on CIFAR-10 and $r=+0.80$ for
VGG-16 on CIFAR-100.  This holds even though $G_T$ is derived for ULA while the
estimate $\widehat{G}_T$ is computed from SGD and the subset models are also retrained with SGD.

\textbf{Gradient Swap matches established selectors.}
\Cref{fig:baseline_comparison} shows test accuracy against retention fraction.
Across datasets and architectures, Gradient Swap performs on par with the static
baselines: its accuracy falls within the spread of the baselines at every
retention fraction, and it is neither consistently ahead of nor behind them. 
This is notable because Gradient Swap is a plain local search. We do not read these
results as an argument for this particular algorithm. They show that a simple
attempt at optimizing the estimate $\widehat{G}_T$ already matches established selectors,
which, together with the correlation above, suggests that $G_T$ is a good target
for developing stronger selection algorithms. Furthermore, the spread of the selectors in \Cref{fig:baseline_comparison} is in line with the DeepCore benchmark
\citep{guo2022deepcorecomprehensivelibrarycoreset}, which concludes that random selection remains a strong baseline. Swap descent can also refine bad selectors: warm-starting it from GraphCut, Herding, Random or EL2N lowers $\widehat{G}_T$ in every case and raises EL2N's accuracy by 2.6 points, while leaving the other three unchanged (Appendix~\ref{app:warm_start}).
\begin{table}[t]
\small
\centering
\caption{Test accuracy (\%) on CIFAR-10 with ResNet-18 as a function of the number
of retained checkpoints $\tau$, at three retention fractions}
\label{tab:checkpointing}
\begin{tabular}{c ccc}
\toprule
$\tau$ & 10\% & 20\% & 40\% \\
\midrule
1  & 69.57 $\pm$ 1.00 & 80.12 $\pm$ 0.80 & 87.02 $\pm$ 0.25 \\
4  & 69.77 $\pm$ 0.85 & 80.37 $\pm$ 0.51 & 87.05 $\pm$ 0.39 \\
10 & 69.81 $\pm$ 1.48 & 80.13 $\pm$ 0.77 & 87.11 $\pm$ 0.34 \\
15 & 69.06 $\pm$ 1.35 & 80.30 $\pm$ 0.80 & 87.05 $\pm$ 0.23 \\
20 & 69.86 $\pm$ 0.72 & 80.34 $\pm$ 0.65 & 86.97 $\pm$ 0.21 \\
\bottomrule
\end{tabular}
\end{table}

%% file: conclusion.tex
\section{Discussion and Limitations}

In this paper, we revisited the theoretical and practical foundations of data selection from first principles. We established a fundamental statistical limitation: even when the training sample contains an accurate subset, no selector can reliably identify one from the training sample and a validation set of comparable size
(\cref{thm:one-sample-oracle-separation}). 
This impossibility result motivated a shift from population risk minimization to \emph{faithfulness}—ensuring subset training closely tracks the full-data parameter distribution. By modeling gradient-based learning via ULA, we bounded the
faithfulness gap by the trajectory gradient discrepancy $G_T(I)$, a training-only
quantity whose off-diagonal structure captures redundancy.
Although minimizing $G_T$ is NP-hard, 
our swap descent algorithm guarantees a decrease of the bound at every step. Empirically, an efficient single-checkpoint implementation matches established static heuristics on CIFAR-10/100 test accuracy.

While our algorithm performs on par with existing methods on average accuracy, this is highly encouraging: (1) it validates that directly optimizing our mathematically principled proxy ($G_T$) is sufficient to match top-tier heuristics, and (2) minimizing $G_T(I)$ guarantees faithfulness across \emph{all} bounded downstream queries (\cref{cor:ula-faithfulness}), providing robustness guarantees where average-case heuristics fail.

Our work opens up many directions for investigation. Our lower bound suggests a deeper connection between data selection and various forms of data memorization \citep{brown2021memorization,attias_arxiv}. Our guarantee is proven for ULA and
is one-sided; a faithfulness bound for SGD or Adam would close the gap between
analysis and practice. $G_T$ is a difference of submodular functions
\citep{narasimhan2012submodularsupermodularprocedureapplicationsdiscriminative,el2023difference},
so stronger solvers apply to the same objective directly, and random projections
(\cref{app:jl_embeddings_ablation}) and proxy models \citep{coleman2019selection}
point to selection at scale.

%% file: appendix.tex
\appendix
\section{Notation}
\label{sec:notation}
Let $\mathcal{Z}$ denote the space of data points and $P \in \mathcal{P}(\mathcal{Z})$ denote the underlying data distribution. We consider the standard learning problem where we have $i.i.d.$ training samples $S = (Z_1, \dots, Z_n) \sim P^n$ from an unknown data distribution $P$. For $I \subseteq [n]$, write $S_I := (Z_i)_{i \in I}$. A data selection procedure with budget $m$ identifies an index set $I \subseteq [n]$ with cardinality $|I| \leq m$. Models are parameterized by a vector $\theta \in \mathbb{R}^d$, and the loss incurred on an individual sample $Z_i$ is given by $\ell(\theta, Z_i)$. The full-dataset empirical risk is defined as $L_S(\theta) = \frac{1}{n} \sum_{i=1}^n \ell(\theta, Z_i)$, while the corresponding subset empirical risk is denoted by $L_{S_I}(\theta) = \frac{1}{|I|} \sum_{i \in I} \ell(\theta, Z_i)$. For a given (randomized) learning algorithm $\Alg$ mapping a dataset to a distribution over $\mathbb{R}^d$, the population risk of a hypothesis $\theta \sim \Alg(S_I)$ trained on $S_I$ is defined as $R_P(\Alg(S_I)) := \mathbb{E}_{Z \sim P,\theta \sim \Alg(S_I)} [\ell(\theta, Z)]$, while the Bayes-optimal population risk is given by $R^\star_P := \inf_f \mathbb{E}_{Z \sim P} [\ell(f, Z)]$, the infimum being over all measurable predictors. For iterative learners, $\mu_t^{S_I}$ denotes the law of the iterate $\theta_t$ when training on $S_I$, and $\Alg(S_I) = \mu_T^{S_I}$ at the final iteration $T$.

\section{Proof of Selection Lower Bound}\label{app:construction-proof}

\subsection{Lower bound}\label{sec:lb}

\subsubsection{Data distribution and meta-distribution}\label{sec:construction}

Recall $\gamma=\tfrac13-\lambda>0$ and $m=\lfloor\lambda n\rfloor$. Throughout, ``for large $n$'' means $n\ge n_\lambda$ for a threshold depending only on $\lambda$.

\begin{figure}[h]
    \centering
    \includegraphics[width=0.85\linewidth]{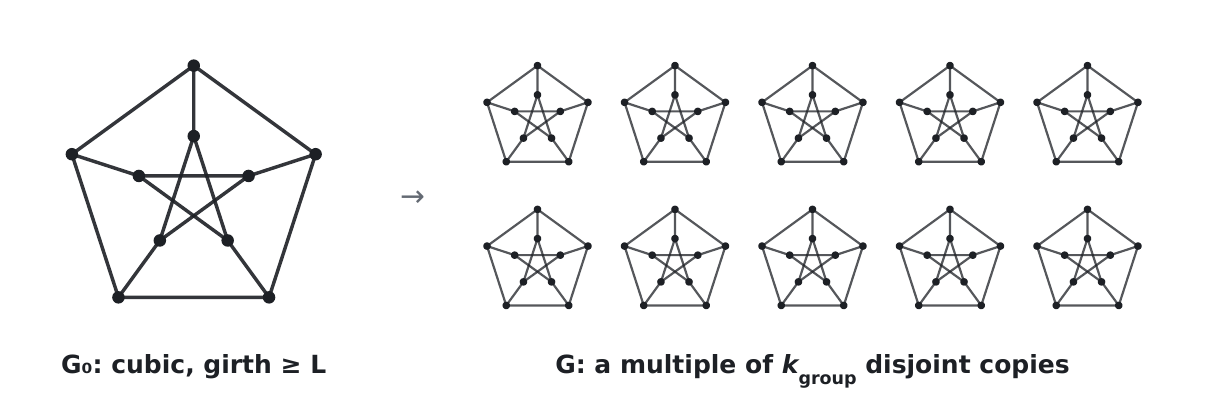}
    \caption{The graph $G$ is a disjoint union of copies of a small cubic graph $G_0$ of large girth. The Petersen graph, which has girth $5$, is shown for illustration; the actual $G_0$ has girth $L\approx\frac12\log_2 n$. Taking a multiple of $\kg$ copies ensures $\kg\mid N$ and $n-\sqrt n<N\le n$.}
    \label{fig:graph}
\end{figure}

The subpopulations of the construction are the edges of a cubic graph whose cycles are long, so that the neighborhood of every edge is a tree. The girth cannot be made arbitrarily large, since a cubic graph of girth $L$ has at least $2^{L/2}$ vertices, but logarithmic girth suffices for our purposes. We use the classical fact that for every integer $L\ge3$ there is a simple cubic graph with girth at least $L$ and fewer than $2^{L+1}$ vertices, where the girth is the length of the shortest cycle. Rather than using a single graph of this kind, we take $G$ to be a disjoint union of copies of a small one, which allows us to control the number of edges precisely; in particular, we make it a multiple of the group size
\[
    \kg:=\Bigl\lceil\frac{10}{\lambda}\Bigr\rceil,
\]
whose value is justified below, when the groups are introduced. Let
\[
    L:=\Bigl\lfloor\log_2\frac{\sqrt n}{3\kg}\Bigr\rfloor,
\]
which is at least $3$ for large $n$, and let $G_0$ be a cubic graph with girth at least $L$ and fewer than $2^{L+1}$ vertices, hence fewer than $3\cdot2^L\le\sqrt n/\kg$ edges. We take $G$ to be the disjoint union of $\kg\bigl\lfloor n/(\kg|E(G_0)|)\bigr\rfloor$ copies of $G_0$ (\cref{fig:graph}), and write $N:=|E(G)|$ for its number of edges. Since $L$ and $\kg$ depend only on $(\lambda,n)$, so does $G$; in particular, $G$ does not depend on $v$.

\begin{lemma}[Properties of $G$]\label{lem:graph}
For large $n$:
\begin{enumerate}[label=(\alph*)]
    \item $G$ is cubic: every vertex has exactly three incident edges. Hence $|V(G)|=2N/3$.
    \item The girth of $G$ is at least $L$, and $L\ge\tfrac12\log_2n-\log_2(6\kg)$.
    \item $n-\sqrt n<N\le n$ and $\kg\mid N$. In particular, $N\ge n/2$.
    \item Every connected set of fewer than $L$ edges of $G$ forms a tree.
\end{enumerate}
\end{lemma}

\begin{proof}
(a) Disjoint copies of a cubic graph form a cubic graph, and counting edge endpoints gives $3|V(G)|=2N$.
(b) A cycle of $G$ lies in a single copy of $G_0$, so it has length at least $L$. The bound on $L$ follows from $\lfloor x\rfloor\ge x-1$.
(c) Write $E_0:=|E(G_0)|$. Then $N=\kg\lfloor n/(\kg E_0)\rfloor E_0$, which lies in $(n-\kg E_0,\,n]$, and $\kg E_0\le\sqrt n$. The number of copies is a multiple of $\kg$, so $\kg\mid N$.
(d) A connected set of edges that contains a cycle has at least as many edges as that cycle, so at least $L$ edges by (b).
\end{proof}

The lower bound uses only parts (a) and (c) of \cref{lem:graph}; part (d), which is a consequence of the girth, is used in the upper bound in \cref{sec:ub}.

\begin{figure}[h]
    \centering
    \begin{minipage}{0.36\linewidth}\centering
        \includegraphics[width=0.85\linewidth]{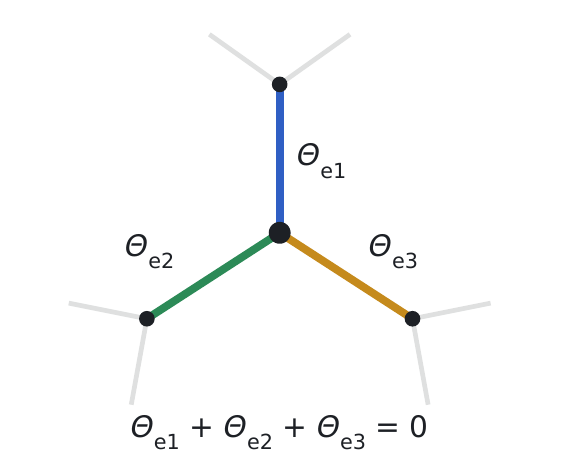}
    \end{minipage}\hfill
    \begin{minipage}{0.4\linewidth}\centering
        \includegraphics[width=0.9\linewidth]{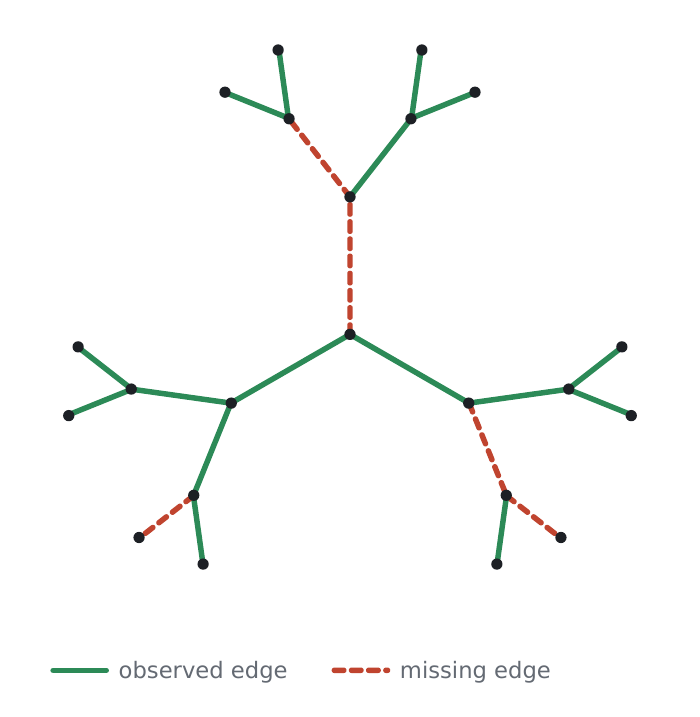}
    \end{minipage}
    \caption{Left: the parity constraint \eqref{eq:parity} at one vertex; over $\F_2$, any two labels at a vertex determine the third. Right: in a graph of girth at least $L$, the ball of radius less than $L/2$ around any vertex is a tree. Edges missing from $S$ (red, dashed) form components surrounded by observed edges (green), and by \cref{lem:graph}(d), components with fewer than $L$ edges are trees.}
    \label{fig:local}
\end{figure}

We now define the labels and the active edges. We fix a partition of $E(G)$ into $J:=N/\kg$ groups $\mathcal G_1,\dots,\mathcal G_J$ of $\kg$ edges each, writing $\mathcal G_j=\{e(j,1),\dots,e(j,\kg)\}$; like $G$, the partition depends only on $(\lambda,n)$ and is known to the learner. Each distribution in the family is indexed by a pair $(\Theta,h)$. The first component, $\Theta=(\Theta_e)_{e\in E(G)}$, assigns to each edge $e$ a label $\Theta_e\in\F_2^B$, subject to the parity constraints
\begin{equation}\label{eq:parity}
    \Theta_{e_1}+\Theta_{e_2}+\Theta_{e_3}=0\quad\text{at every vertex, where $e_1,e_2,e_3$ are its three edges}
\end{equation}
(\cref{fig:local}, left). We identify the label set $[\ncls]$ with $\F_2^B$, so that there are $\ncls:=2^B$ classes and an incorrect $B$-bit string counts as a single mistake. The label length $B$ governs the comparison, made in the lower bound, between the number of bits carried by the selected sample and the number of bits needed to describe the labels. Besides $mB$ bits of labels, the selected sample conveys up to $n$ bits through the choice of the index set, and the argument spends up to $n$ further bits on locating the mistakes of the learner. These overheads do not grow with $B$, whereas the entropy of the labels grows linearly in $B$, so they become negligible when $B$ is large compared with $1/\gamma$. At the same time, $B$ must not depend on $n$, so that the number of classes $\ncls=2^B$ depends only on $\lambda$. We take
\[
    B:=\Bigl\lceil\frac{8}{\gamma}\Bigr\rceil.
\]
The second component, $h=(h_1,\dots,h_J)\in[\kg]^J$, selects one \emph{active edge} $e(j,h_j)$ in each group; the risk will be incurred only on active edges, so that an oracle that knows $h$ needs information about a single edge per group. Choosing $\kg=\lceil10/\lambda\rceil$ ensures that the $J=N/\kg$ active edges require at most $\frac35\lambda n<m$ certificate examples in expectation (\cref{lem:budget}), while losing only a factor of $1/\kg=\Theta(\lambda)$ when converting prediction errors across a group into test risk. Throughout, the edge $e$ plays the role of a subpopulation identity, which is part of every input, whereas its label $\Theta_e$ is unknown.

We next describe the distribution $P_{\Theta,h}$ of a single example. An input is a pair $(e,z)$ with $e\in E(G)$ and $z\in\F_2^B\cup\{\bot\}$. Given the masking probability $\pmask\in(0,1)$, specified below, an example $Z=(X,Y)\sim P_{\Theta,h}$ is drawn as
\[
    Z=(X,Y)=\begin{cases}
        \bigl((e,\Theta_e),\ \Theta_e\bigr), & e\sim\mathrm{Unif}(E(G)),\ \ \text{w.p. } 1-\pmask\ \ \text{(clear)},\\[3pt]
        \bigl((e(j,h_j),\bot),\ \Theta_{e(j,h_j)}\bigr), & j\sim\mathrm{Unif}([J]),\ \ \text{w.p. } \pmask\ \ \text{(masked)}.
    \end{cases}
\]
Since the input of a \emph{clear} example contains its label, a learner can predict correctly on clear inputs by copying $z$, and the risk is therefore determined by the \emph{masked} inputs, which present an active edge with $\bot$ in place of its label. The two kinds of examples play different roles in $(S,V)$. Clear examples fall on uniformly random edges, independently of $h$, and thus reveal labels but carry no information about the active edges. A masked example, in contrast, reveals the active edge of its group, since only active edges are masked. As explained above, $\pmask$ should be large, because it multiplies the risk, but small enough that most groups receive no masked example in $(S,V)$. We take
\[
    \pmask:=\frac{\gamma}{8\kg}\cdot\frac{n}{n+v},
\]
so that the expected number of masked examples in $(S,V)$ is $(n+v)\pmask=\gamma n/(8\kg)$, which is at most a $\gamma/4$ fraction of the $J\ge n/(2\kg)$ groups, irrespective of the validation size $v$. Since $\kg\ge31$ and $\gamma<\tfrac13$, we also have
\begin{equation}\label{eq:pmask}
    \pmask\ \le\ \frac{\gamma}{8\kg}\ <\ \frac{1}{720}.
\end{equation}
Training, validation and test examples are i.i.d.\ from $P_{\Theta,h}$, and the family $\Pfam$ consists of all such $P_{\Theta,h}$. It is finite, its number of classes $\ncls$ depends only on $\lambda$, and $R^\star_P=0$ because the label is a function of the input.

For the lower bound, we place a prior on the family, the \emph{meta-distribution}, under which $(\Theta,h)$ is uniform over all valid pairs. To describe it, note that for a fixed bit position $k$, the bit-plane $(\Theta_e[k])_{e\in E(G)}$ satisfies \eqref{eq:parity} exactly when it lies in the \emph{cycle space}
\[
    \mathcal C(G):=\bigl\{x\in\F_2^{E(G)}:\ x_{e_1}+x_{e_2}+x_{e_3}=0\ \text{at every vertex}\bigr\}.
\]
This is the solution set of $|V(G)|$ homogeneous linear equations in $N$ unknowns, so $\dim\mathcal C(G)\ge N-|V(G)|=N/3$ by \cref{lem:graph}(a). Under the meta-distribution, $\Theta$ and $h$ are independent, $h$ is uniform on $[\kg]^J$, and the $B$ bit-planes of $\Theta$ are i.i.d.\ uniform on $\mathcal C(G)$. Although the parity constraints make the labels dependent, $\Theta$ is uniform over $|\mathcal C(G)|^B=2^{B\dim\mathcal C(G)}$ values, and hence
\begin{equation}\label{eq:entropy}
    H(\Theta)=B\cdot\dim\mathcal C(G)\ \ge\ \frac{NB}{3}\qquad\text{(in bits)}.
\end{equation}
Since the family is finite, it suffices to show that the risk averaged over this prior is large.

\subsubsection{Main argument}\label{sec:lb-main}

Fix a learner $\Alg_m$ and a selector $\Pi$. Draw $(\Theta,h)$ from the meta-distribution, then $S\sim P_{\Theta,h}^n$ and $V\sim P_{\Theta,h}^v$, with independent internal randomness for $\Pi$ and $\Alg_m$, and let $\hat f:=\Alg_m(S_{\widehat I_m})$. In this subsection, $\Pr$, $\E$, $H$ and $I$ refer to all of this randomness, and our goal is
\begin{equation}\label{eq:goal}
    \E\bigl[R_{P_{\Theta,h}}(\hat f)\bigr]\ \ge\ \frac{\lambda\gamma^2}{264}\cdot\frac{n}{n+v}.
\end{equation}

The argument conditions on everything except the labels. Let $W$ consist of the edges $e_1,\dots,e_{n+v}$ of all examples in $(S,V)$, the indicators of which of these examples are masked, and the internal randomness of $\Pi$ and $\Alg_m$. We emphasize that $W$ excludes the inputs themselves, since the input of a clear example contains its label.

\begin{lemma}[Properties of $W$]\label{lem:W}
\leavevmode
\begin{enumerate}[label=(\roman*)]
    \item $W$ is independent of $\Theta$ (but not of $h$).
    \item $(S,V)$, $\widehat I_m$ and $\hat f$ are functions of $(W,\Theta)$.
    \item Let $M:=\bigl(\widehat I_m,\ (Y_i)_{i\in\widehat I_m}\bigr)$. Given $W$, the selected sample $S_{\widehat I_m}$ is a function of $M$; hence $\hat f$ is a function of $(W,M)$.
\end{enumerate}
\end{lemma}

\begin{proof}
(i) The edges and the mask indicators are drawn independently of $\Theta$; the masked edges depend on $h$.
(ii) Given $W$, example $i$ equals $((e_i,\Theta_{e_i}),\Theta_{e_i})$ if it is clear and $((e_i,\bot),\Theta_{e_i})$ if it is masked, and the selector and the learner are deterministic given their internal randomness, which is part of $W$.
(iii) For $i\in\widehat I_m$, $W$ provides $e_i$ and the mask indicator, so $Z_i$ is recovered from $(i,Y_i)$: $X_i=(e_i,Y_i)$ if example $i$ is clear and $X_i=(e_i,\bot)$ if it is masked.
\end{proof}

A masked example in $(S,V)$ reveals the active edge of its group, so the selector can be uncertain about $h_j$ only if no masked example from $\mathcal G_j$ occurs. We call a group $\mathcal G_j$ \emph{resolved} if some example in $(S,V)$ is a masked example from $\mathcal G_j$, and \emph{unresolved} otherwise. Let $U\subseteq[J]$ be the set of unresolved groups and $E_U:=\bigcup_{j\in U}\mathcal G_j$; both are functions of $W$. We count the mistakes of $\hat f$ on all edges of unresolved groups,
\[
    Q:=\bigl|\{e\in E_U:\ \hat f(e,\bot)\ne\Theta_e\}\bigr|.
\]
The edges of resolved groups are excluded from this count, and the choice of $\pmask$ ensures that they form a small fraction of all edges, namely at most
\[
    \pres:=\frac{\gamma}{4}
\]
in expectation, a quarter of the gap $\gamma$.

\begin{lemma}[Few groups are resolved]\label{lem:resolved}
For large $n$ and every $j$, $\Pr(j\notin U)\le\pres$. Hence $\E|E(G)\setminus E_U|\le\pres N$.
\end{lemma}

\begin{proof}
A single example is a masked example from $\mathcal G_j$ with probability $\pmask/J$. A union bound over the $n+v$ examples, together with $(n+v)\pmask=\gamma n/(8\kg)$, $J=N/\kg$ and $N\ge n/2$, gives
\[
    \Pr(j\notin U)\ \le\ (n+v)\frac{\pmask}{J}\ =\ \frac{\gamma}{8}\cdot\frac nN\ \le\ \frac{\gamma}{4}=\pres.
\]
Finally, $\E|E(G)\setminus E_U|=\kg\sum_j\Pr(j\notin U)\le \kg J\pres=\pres N$.
\end{proof}

Although the risk involves only the active edge of each group, every mistake counted by $Q$ contributes to it, because the active edge of an unresolved group is uniformly distributed given everything on which $\hat f$ depends.

\begin{lemma}[Risk lemma]\label{lem:risk}
$\E\bigl[R_{P_{\Theta,h}}(\hat f)\bigr]\ \ge\ \dfrac{\pmask}{n}\,\E[Q]$.
\end{lemma}

\begin{proof}
We first show that $h_j$ is uniform on $[\kg]$ given $(W,\Theta)$ for every $j\in U$. Given $(\Theta,h)$, the law of $W$ does not depend on $\Theta$, and it depends on $h_j$ only through the masked examples from $\mathcal G_j$, of which there are none if $j\in U$. Since $h$ is uniform and independent of $\Theta$, the claim follows from Bayes' rule, while $\hat f$ is determined by $(W,\Theta)$ by \cref{lem:W}(ii).

Restricting the risk to the masked inputs of unresolved groups, each of which has probability $\pmask/J$, gives
\[
    R_{P_{\Theta,h}}(\hat f)\ \ge\ \sum_{j\in U}\frac{\pmask}{J}\,\ind\bigl\{\hat f(e(j,h_j),\bot)\ne\Theta_{e(j,h_j)}\bigr\}.
\]
Averaging over the uniform $h_j$, conditionally on $(W,\Theta)$, turns the $j$-th term into
\[
    \frac{\pmask}{J\kg}\sum_{\ell=1}^{\kg}\ind\bigl\{\hat f(e(j,\ell),\bot)\ne\Theta_{e(j,\ell)}\bigr\}.
\]
Summing over $j\in U$ and using $J\kg=N\le n$,
\[
    \E\bigl[R_{P_{\Theta,h}}(\hat f)\,\big|\,W,\Theta\bigr]\ \ge\ \frac{\pmask}{N}\,Q\ \ge\ \frac{\pmask}{n}\,Q,
\]
and taking expectations proves the lemma.
\end{proof}

It remains to show that $\E[Q]$ is of order $\gamma n$, and it is at this step that the restriction to $m$ selected examples enters. By \cref{lem:W}, conditionally on $W$ we have the Markov chain
\begin{equation}\label{eq:markov}
    \Theta\ \longrightarrow\ (S,V)\ \longrightarrow\ M=\bigl(\widehat I_m,(Y_i)_{i\in\widehat I_m}\bigr)\ \longrightarrow\ \hat f,
\end{equation}
in which $(S,V)$ is a function of $\Theta$, $M$ is computed from $(S,V)$ by the selector, and $\hat f=F(W,M)$ for a fixed function $F$. Although the selector may use every label in $(S,V)$, the learner has access to these labels only through $M$, and the next lemma quantifies the resulting loss.

\begin{lemma}[Bits lemma]\label{lem:bits}
For $n\ge n_\lambda$, it holds that $\displaystyle\E[Q]\ \ge\ \Bigl(\frac13-\pres\Bigr)N-m-\frac{2n}{B}$.
\end{lemma}

\begin{proof}
Recall that $M=\bigl(\widehat I_m,(Y_i)_{i\in\widehat I_m}\bigr)$. The proof is based on comparing the conditional entropy $H(\Theta\mid W)$ from above and below. Since $\widehat I_m\subseteq[n]$ takes at most $2^n$ values and, given $\widehat I_m$, the labels $(Y_i)_{i\in\widehat I_m}\in(\F_2^B)^{|\widehat I_m|}$ take at most $2^{mB}$ values, we have
\begin{equation}\label{eq:HM}
    H(M\mid W)\ \le\ n+mB.
\end{equation}
The contribution of $\widehat I_m$ cannot be omitted, since a selector that observes all labels can encode information about the unselected labels in its choice of indices; the choice of $B$ makes it negligible compared with $NB$.

Next, we upper bound $H(\Theta\mid W,\hat f)$. Let $\chi\in\{0,1\}^{E(G)}$ be the error indicator vector defined by $\chi_e=\ind\{e\in E_U\text{ and }\hat f(e,\bot)\ne\Theta_e\}$, so that $\|\chi\|_0=Q$. Notice that, conditioned on $(W,\hat f,\chi)$, the label $\Theta_e$ is determined by $\hat f(e,\bot)$ for every $e\in E_U$ with $\chi_e=0$; hence $\Theta$ is uniquely determined by specifying the $Q$ labels $\{\Theta_e:\ e\in E_U,\ \chi_e=1\}$ together with the $|E(G)\setminus E_U|$ labels on the resolved groups, $\Theta_{E(G)\setminus E_U}$, where $E_U$ is a function of $W$. Since $\chi$ takes at most $2^N$ values and each label is a $B$-bit string, the chain rule of entropy gives
\begin{equation}\label{eq:fano}
\begin{aligned}
    H(\Theta\mid W,\hat f)\ &\le\ H(\chi\mid W,\hat f)+H(\Theta\mid W,\hat f,\chi)\\
    &\le\ N+B\,\E[Q]+B\,\E|E(G)\setminus E_U|\ \le\ n+B\,\E[Q]+\pres NB,
\end{aligned}
\end{equation}
where the last step uses \cref{lem:resolved} and $N\le n$. Inequality \eqref{eq:fano} is a coordinatewise analogue of Fano's inequality $H(\Theta\mid\hat\Theta)\le h(P_e)+P_e\log(|\mathcal X|-1)$: the term $N$, which accounts for the error pattern, corresponds to $h(P_e)$, the term $B\,\E[Q]$ corresponds to $P_e\log(|\mathcal X|-1)$, and the term $\pres NB$ accounts for the resolved groups, on which mistakes are not counted.

Finally, we compare the two bounds. Since $\Theta$ is independent of $W$ by \cref{lem:W}(i), \eqref{eq:entropy} gives $H(\Theta\mid W)=H(\Theta)\ge NB/3$. Conversely, since $\hat f$ is a deterministic function of $(W,M)$ by \cref{lem:W}(iii), the chain rule and the fact that conditioning reduces entropy give
\[
    H(\Theta\mid W)\  \le\ H(M\mid W)+H(\Theta\mid W,\hat f).
\]
Combining these two bounds with \eqref{eq:HM} and \eqref{eq:fano}, we obtain
\[
    \frac{NB}{3}\ \le\ (n+mB)+\bigl(n+B\,\E[Q]+\pres NB\bigr),
\]
and dividing by $B$ proves the lemma.
\end{proof}

We can now complete the proof of the lower bound.

\begin{proof}[Proof of \cref{thm:one-sample-oracle-separation} Lower Bound]
By \cref{lem:graph}(c), $N>n-\sqrt n$. Moreover, $m\le\lambda n$, $\pres=\gamma/4$, and $2/B\le\gamma/4$ since $B\ge8/\gamma$. Hence \cref{lem:bits} gives, for large $n$,
\begin{align*}
    \E[Q]\ &\ge\ \Bigl(\frac13-\lambda-\pres-\frac2B\Bigr)n-\sqrt n
     \\
    &\ge \frac{\gamma}{2}n-\sqrt n\ \\
    &\ge \frac{\gamma n}{3}.
\end{align*}
By \cref{lem:risk} and $\kg\le10/\lambda+1\le11/\lambda$,
\[
    \E\bigl[R_{P_{\Theta,h}}(\hat f)\bigr]\ \ge\ \frac{\pmask}{n}\cdot\frac{\gamma n}{3}
    \ =\ \frac{\gamma^2}{24\kg}\cdot\frac{n}{n+v}
    \ \ge\ \frac{\lambda\gamma^2}{264}\cdot\frac{n}{n+v},
\]
which is \eqref{eq:goal}. The left-hand side is the average over the meta-distribution of $\E_{S,V,\Pi,\Alg_m}[R_P(\hat f)]$ with $P=P_{\Theta,h}$, so some $P\in\Pfam$ satisfies the bound. This $P$ may depend on $(\Alg_m,\Pi)$, whereas the family does not.
\end{proof}

\subsection{Upper bound}\label{sec:ub}

We now fix an arbitrary member $P=P_{\Theta,h}$ of $\Pfam$. Probabilities in this subsection are over $S\sim P^n$ only, and all bounds are uniform over the family.

The learner $\Astar$ knows the graph $G$ and the label length $B$, which depend only on $(\lambda,n)$, but not $(\Theta,h)$. Given a training sample, it discards the masked examples and forms a linear system over $\F_2$ in the unknowns $\Theta$, consisting of
\[
    \underbrace{\Theta_{e_1}+\Theta_{e_2}+\Theta_{e_3}=0\ \text{ at every vertex}}_{\text{parity \eqref{eq:parity}}}
    \quad\text{and}\quad
    \underbrace{\Theta_e=z\ \text{ for each clear }(e,z)\text{ in the sample}}_{\text{observations}}.
\]
It computes one solution by Gaussian elimination with a fixed pivoting rule, setting free variables to zero (if there is no solution, it takes $\Theta=0$). On a clear input $(e,z)$ it outputs $z$, and on a masked input $(e,\bot)$ it outputs the value of $\Theta_e$ in the computed solution. The analysis of $\Astar$ rests on the following two observations.

\begin{lemma}[Two basic facts]\label{lem:basic}
\leavevmode
\begin{enumerate}[label=(\roman*)]
    \item For every $I\subseteq[n]$, $R_P(\Astar(S_I))\le\pmask$.
    \item Say that $S_I$ \emph{determines} $\Theta_e$ if all solutions of its linear system agree on $\Theta_e$. If $S_I$ determines $\Theta_e$, then $\Astar(S_I)$ is correct on $(e,\bot)$, and so is $\Astar(S_{I'})$ for every $I'\supseteq I$.
\end{enumerate}
\end{lemma}

\begin{proof}
(i) $\Astar$ copies the displayed label, so it is correct on every clear input, and masked inputs have total probability $\pmask$.
(ii) The true $\Theta$ satisfies every equation obtained from $S$, so it is one of the solutions, and all solutions agree on $\Theta_e$. A larger subsample only adds equations that the true $\Theta$ satisfies, which cannot enlarge the solution set.
\end{proof}

By \cref{lem:basic}, both upper bounds reduce to exhibiting, for each active edge, a small set of examples that determines its label. For an edge that appears in $S$, a single example suffices, and the other edges require the parity constraints. We call an edge \emph{observed} if $S$ contains a clear example of it, and \emph{missing} otherwise. For a missing edge $e$, let $T_e$ be the connected component containing $e$ in the subgraph of missing edges, with $|T_e|$ edges (and $T_e:=\emptyset$ for an observed edge), and let $\partial T_e$ be the set of edges outside $T_e$ that share a vertex with $T_e$. The boundary edges are all observed, since a missing edge sharing a vertex with $T_e$ would belong to $T_e$.

We truncate at component size
\[
    s:=\lfloor L/2\rfloor,\qquad\text{so that}\qquad \tfrac18\log_2 n\le s\le\log_2 n\ \ \text{for large $n$}
\]
by \cref{lem:graph}(b). Since $s<L$, \cref{lem:graph}(d) shows that every component with at most $s$ edges is a tree; meanwhile, $s$ is large enough that larger components occur with probability $O(1/\log n)$ (\cref{lem:cert}(iii)), yet small enough ($s\ll\sqrt n$) to control the variance of the oracle's budget (\cref{lem:budget}).

The \emph{certificate} of an edge $e$ is the set $I_e\subseteq[n]$ consisting of the smallest index of a clear example of $e$ if $e$ is observed; of the smallest index of a clear example of each edge of $\partial T_e$ if $e$ is missing and $|T_e|\le s$; and $I_e:=\emptyset$ if $e$ is missing and $|T_e|>s$, in which case we say that the certificate \emph{fails}. The use of smallest indices makes $I_e$ a canonical subset of $[n]$ determined by $S$, which the oracle retains in the proof of \cref{thm:one-sample-oracle-separation}; \cref{lem:budget} does not depend on this choice, since it involves only the set of observed edges.

\begin{figure}[h]
    \centering
    \includegraphics[width=0.3\linewidth]{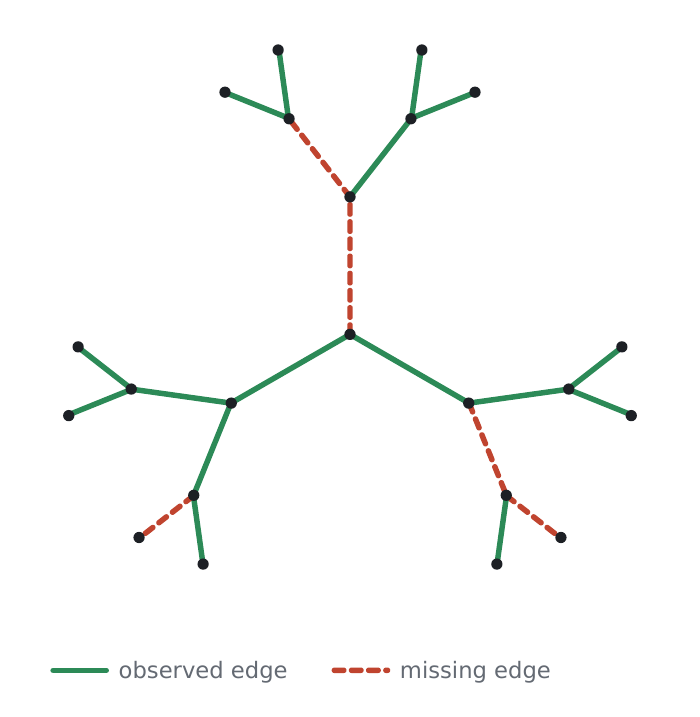}
    \caption{A missing edge $e$ (red) in a component $T_e$ of missing edges (dashed), surrounded by the observed boundary edges $\partial T_e$ (green). Removing $e$ splits the tree into two parts; summing the parity equations over the vertex set $V_1$ of one part cancels every edge inside $V_1$ and expresses $\Theta_e$ as the sum of the labels of the boundary edges leaving $V_1$.}
    \label{fig:tree}
\end{figure}

\begin{lemma}[Certificates]\label{lem:cert}
For every edge $e$ and every $P\in\Pfam$:
\begin{enumerate}[label=(\roman*)]
    \item If the certificate of $e$ does not fail, then every subsample of $S$ containing $S_{I_e}$ determines $\Theta_e$.
    \item $|I_e|\le|T_e|+3$, $|I_e|\le s+3$, and $\E|I_e|\le6$.
    \item The certificate of $e$ fails with probability at most $3/s$.
\end{enumerate}
\end{lemma}

\begin{figure}[h]
    \centering
    \includegraphics[width=0.3\linewidth]{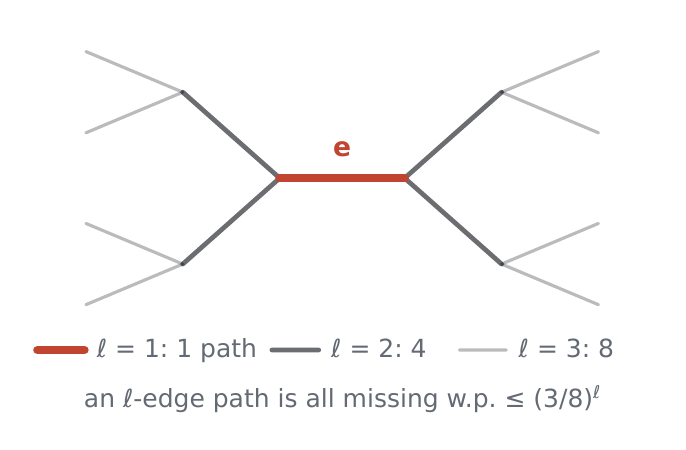}
    \caption{Simple paths starting with $e$ in a cubic graph: there are at most $2^\ell$ of length $\ell$, and each consists of missing edges with probability at most $(3/8)^\ell$. Since $2\cdot\frac38<1$, the expected size of the component $T_e$ is bounded, as for a subcritical branching process.}
    \label{fig:branch}
\end{figure}

\begin{proof}
(i) If $e$ is observed, the subsample contains the equation $\Theta_e=z$. Otherwise $T_e$ is a tree, and removing $e$ splits it into two trees (\cref{fig:tree}). Let $V_1$ be the vertex set of one of them (possibly a single vertex), and add up the parity equations at the vertices of $V_1$. Every edge with both endpoints in $V_1$ appears twice and cancels, so the labels of the edges with exactly one endpoint in $V_1$ sum to zero. One of these edges is $e$; every other edge of $T_e$ has either both or no endpoints in $V_1$, so the remaining edges in the sum lie in $\partial T_e$, and the subsample contains a clear example of each of them. Hence
\[
    \Theta_e\ =\ \sum_{f\in\partial T_e:\ |f\cap V_1|=1}\Theta_f
\]
is determined.

For (ii) and (iii), we first bound the probability that given edges are missing. A training example is a clear example of a fixed edge with probability $(1-\pmask)/N$. Since $N\le n$ (\cref{lem:graph}(c)) and $\pmask<1/720$ by \eqref{eq:pmask}, for any $k$ distinct edges,
\begin{equation}\label{eq:missing}
    \Pr(\text{all $k$ edges are missing})\ =\ \Bigl(1-\frac{k(1-\pmask)}{N}\Bigr)^n\ \le\ e^{-k(1-\pmask)}\ \le\ \Bigl(\frac38\Bigr)^k,
\end{equation}
because $e^{-719/720}\approx0.3684<\frac38$. Next, every edge of $T_e$ is the last edge of a simple path in $G$ that starts with $e$ and consists of missing edges: take a shortest chain of edges of $T_e$ from $e$ to it in which consecutive edges share a vertex; by minimality, non-consecutive edges of the chain share no vertex, so the chain is a simple path. Since $G$ is cubic, for every $\ell\ge1$ at most $2^\ell$ simple paths with $\ell$ edges start with $e$ (\cref{fig:branch}). Applying \eqref{eq:missing} to the $\ell$ edges of each path gives
\begin{equation}\label{eq:component}
    \E|T_e|\ \le\ \sum_{\ell\ge1}2^\ell\Bigl(\frac38\Bigr)^\ell\ =\ \sum_{\ell\ge1}\Bigl(\frac34\Bigr)^\ell\ =\ 3.
\end{equation}

(ii) If $e$ is missing and $|T_e|\le s$, then $T_e$ is a tree with $|T_e|+1$ vertices, each of degree three in $G$, and its edges use $2|T_e|$ of the $3(|T_e|+1)$ edge--vertex incidences at these vertices. Hence $|I_e|=|\partial T_e|\le|T_e|+3\le s+3$. If $e$ is observed, then $|I_e|=1$ and $|T_e|=0$; if the certificate fails, then $|I_e|=0$. Thus $|I_e|\le|T_e|+3$ and $|I_e|\le s+3$ in every case, and by \eqref{eq:component}, $\E|I_e|\le\E|T_e|+3\le6$.

(iii) The certificate fails only if $|T_e|>s$, which by Markov's inequality and \eqref{eq:component} has probability at most $\E|T_e|/s\le3/s$.
\end{proof}

Since the full sample contains every certificate follows directly.

\begin{proof}[Proof of \cref{thm:one-sample-oracle-separation}]
The full sample $S$ contains $S_{I_e}$ for every edge $e$. By \cref{lem:cert,lem:basic}, $\Astar(S)$ is correct on $(e,\bot)$ unless the certificate of $e$ fails, which has probability at most $3/s$, and the masked inputs have total probability $\pmask$. Therefore
\begin{align*}
    \E_S\bigl[R_P(\Astar(S))\bigr]\ &\le\ \frac{3\,\pmask}{s}\ \le\ \frac{24}{\log_2n}\cdot\frac{\gamma}{8\kg}\cdot\frac{n}{n+v}\\
    &=\ \frac{3\gamma}{\kg}\cdot\frac{n}{(n+v)\log_2n}\ \le\ \frac25\,\lambda\gamma\cdot\frac{n}{(n+v)\log_2n},
\end{align*}
using $s\ge\frac18\log_2n$, $\pmask=\frac{\gamma}{8\kg}\cdot\frac{n}{n+v}$, and $\kg\ge10/\lambda$.
\end{proof}

For part (ii), the oracle is restricted to $m$ examples, but since it may depend on $h$, it needs only the certificates of the $J$ active edges. Distinct active edges may share their certificate: all edges of a missing component $T$ with $|T|\le s$ have the same certificate, consisting of one index for each edge of $\partial T$, so the union of the certificates contains these indices only once per component. We therefore consider
\[
    F_h\ :=\ \bigl|\{j:\ e(j,h_j)\text{ is observed}\}\bigr|\ +\sum_{\substack{T:\ |T|\le s,\\ T\text{ contains an active edge}}}|\partial T|,
\]
where the sum runs over components of missing edges. It satisfies
\begin{equation}\label{eq:Fh}
    \Bigl|\bigcup_{j=1}^J I_{e(j,h_j)}\Bigr|\ \le\ F_h\ \le\ \sum_{j=1}^J\bigl|I_{e(j,h_j)}\bigr|,
\end{equation}
where the first inequality holds because the union consists of one index for each observed active edge and the boundary indices of the components that contain an active edge, and the second because a component containing $r\ge1$ active edges contributes $|\partial T|$ to $F_h$ and $r|\partial T|$ to the sum. The next lemma shows that $F_h\le m$ with high probability.

\begin{lemma}[Budget]\label{lem:budget}
For large $n$, $\E F_h\le\frac35\lambda n$ and $\Pr(F_h>m)\le1458\,s^2/(\lambda^2n)$.
\end{lemma}

\begin{proof}
By \eqref{eq:Fh}, \cref{lem:cert}(ii), $J=N/\kg\le n/\kg$ and $\kg\ge10/\lambda$,
\[
    \E F_h\ \le\ \sum_{j=1}^J\E\bigl|I_{e(j,h_j)}\bigr|\ \le\ 6J\ \le\ \frac{6n}{\kg}\ \le\ \frac35\lambda n,
\]
and since $m\ge\lambda n-1$, we obtain $m-\E F_h\ge\frac25\lambda n-1\ge\lambda n/3$ for large $n$.

To show that $F_h$ concentrates, we bound the effect of replacing a single example and apply the Efron--Stein inequality. The quantity $F_h$ depends on $S$ only through the set of observed edges. Suppose that the status of a single edge $f$ changes, and call a component of missing edges, before or after the change, \emph{affected} if it shares a vertex with $f$. Unaffected components retain their edges, their boundary and their active edges, so their terms do not change. There are at most two affected components before the change and at most two after it: if $f$ is missing, only its own component is affected, and if $f$ is observed, at most one component meets each of its two endpoints. Each component term lies in $[0,s+3]$, since $|\partial T|\le s+3$, so the sum changes by at most $2(s+3)$, the first term by at most one, and $F_h$ by at most $2(s+3)+1$. Replacing one example of $S$ by an independent copy changes the status of at most two edges, namely those of the old and the new example, and therefore changes $F_h$ by at most $2(2s+7)\le18s$.

Writing $F_h^{(i)}$ for the value of $F_h$ after the $i$-th example is replaced, the Efron--Stein inequality gives
\[
    \Var(F_h)\ \le\ \frac12\sum_{i=1}^n\E\bigl[(F_h-F_h^{(i)})^2\bigr]\ \le\ \frac n2(18s)^2\ =\ 162\,n\,s^2,
\]
and Chebyshev's inequality gives
\[
    \Pr(F_h>m)\ \le\ \Pr\Bigl(F_h-\E F_h>\frac{\lambda n}{3}\Bigr)\ \le\ \frac{162\,n\,s^2}{(\lambda n/3)^2}\ =\ \frac{1458\,s^2}{\lambda^2n}.\qedhere
\]
\end{proof}

We now combine \cref{lem:cert,lem:budget} to prove part (ii).

\begin{proof}[Proof of \cref{thm:one-sample-oracle-separation}]
Since $I^\star_m$ minimizes $R_P(\Astar(S_I))$ over subsets of size $m$, it suffices to exhibit a subset $\widetilde I\subseteq[n]$ of size $m$, which may depend on $P$ as $I^\star_m$ does, and to bound $\E_S[R_P(\Astar(S_{\widetilde I}))]$. Call $S$ \emph{good} if $F_h\le m$. If $S$ is good, let $\widetilde I$ contain the union $\bigcup_jI_{e(j,h_j)}$ of the certificates of the active edges, which by \eqref{eq:Fh} has at most $F_h\le m$ indices, together with arbitrary further indices to reach exactly $m$. If $S$ is not good, let $\widetilde I:=[m]$.

If $S$ is good, $S_{\widetilde I}$ contains the certificate of every active edge, so by \cref{lem:cert,lem:basic} it determines the label of every active edge whose certificate does not fail. Hence $\Astar(S_{\widetilde I})$ errs only on inputs $(e(j,h_j),\bot)$ whose certificate fails, each of probability $\pmask/J$. If $S$ is not good, the risk is still at most $\pmask$ by \cref{lem:basic}(i). By \cref{lem:cert}(iii) and \cref{lem:budget},
\[
    \E_S\bigl[R_P(\Astar(S_{I^\star_m}))\bigr]
    \ \le\ \E_S\bigl[R_P(\Astar(S_{\widetilde I}))\bigr]
    \ \le\ \pmask\Bigl(\frac3s+\frac{1458\,s^2}{\lambda^2n}\Bigr).
\]
Since $\frac18\log_2n\le s\le\log_2n$ and $\frac{1458(\log_2n)^2}{\lambda^2n}\le\frac{1}{\log_2n}$ for large $n$, the bracket is at most $\frac{25}{\log_2n}$. Since $\pmask=\frac{\gamma}{8\kg}\cdot\frac{n}{n+v}$,
\[
    \E_S\bigl[R_P(\Astar(S_{I^\star_m}))\bigr]\ \le\ \frac{25\gamma}{8\kg}\cdot\frac{n}{(n+v)\log_2n}\ \le\ \frac25\,\lambda\gamma\cdot\frac{n}{(n+v)\log_2n},
\]
using $\kg\ge10/\lambda$ in the last step. The constants and the threshold $n_\lambda$ depend only on $\lambda$, which completes the proof of \cref{thm:one-sample-oracle-separation}.
\end{proof}

\section{KL divergence in parameter space upper bounds the faithfulness error}\label{app:kl-control}
\begin{proof}
Let $P=\Alg_n(S)$ and $Q=\Alg_m(S_I)$, and write
$\delta=\mathbb E_P[q]-\mathbb E_Q[q]$. If
$D_{\mathrm{KL}}(P\|Q)=\infty$, the claim is immediate. Otherwise, set
$c=\frac{a+b}{2}$. Since $P$ and $Q$ are both probability measures, $(P-Q)(\mathbb R^d)=0$, so
\[
    \delta
    =
    \int \bigl(q-c\bigr)\,\mathrm d(P-Q).
\]
Because $q$ takes values in $[a,b]$, we have $|q-c|\le \frac{b-a}{2}$ pointwise, and therefore
\[
    |\delta|
    \le
    \frac{b-a}{2}\,\|P-Q\|_{1}
    =
    (b-a)\,\mathrm{TV}(P,Q),
\]
where $\mathrm{TV}(P,Q)=\sup_{A}|P(A)-Q(A)|=\frac12\|P-Q\|_1$ is the total variation distance. Pinsker's inequality gives
\[
    \mathrm{TV}(P,Q)\le\sqrt{\tfrac12 D_{\mathrm{KL}}(P\|Q)}.
\]
Combining the two bounds yields
\[
    |\delta|
    \le
    (b-a)\sqrt{\tfrac12 D_{\mathrm{KL}}(P\|Q)}
    =
    \sqrt{\frac{(b-a)^2}{2}\,D_{\mathrm{KL}}(P\|Q)}.
\]
The argument applies directly to $|\delta|$, so no separate treatment of the positive and negative parts is needed. This proof applies to every measurable $q$ bounded in $[a,b]$.
\end{proof}

\section{Proof of the gradient trajectory upper bounding the faithfulness}\label{app:path-kl}

\begin{proof}
For a sample $S'\in\{S,S_I\}$, let $L_{S'}$ be its empirical loss, and let
$(\theta_t^{S'})_{t=0}^{T}$ be the ULA chain
\[
    \theta_{t+1}^{S'} = \theta_t^{S'} - h_t\nabla L_{S'}(\theta_t^{S'})
    + \sqrt{2h_t/\beta_t}\,\xi_{t+1},
    \qquad \xi_{t+1}\stackrel{\mathrm{i.i.d.}}{\sim}\mathcal{N}(0,I_d),
\]
started from the same initial law $\mu_0$ for both samples.
Write $\mu_t^{S'}$ for the law of $\theta_t^{S'}$, so that
$\Alg_n(S)=\mu_T^{S}$ and $\Alg_m(S_I)=\mu_T^{S_I}$, and let
$\mathbb{P}^{S'}$ be the law of the path $(\theta_0^{S'},\dots,\theta_T^{S'})$.
Recall that
\[
    G_T(I) = \sum_{t=0}^{T-1}\frac{\beta_t h_t}{4}\,
    \mathbb{E}_{\theta_t\sim\mu_t^{S}}
    \bigl\|\nabla L_{S_I}(\theta_t)-\nabla L_S(\theta_t)\bigr\|^2 .
\]

Since $\ell(\cdot,z)$ is continuously differentiable, the map
$\theta\mapsto\nabla L_{S'}(\theta)$ is continuous, so the transition kernels
\[
    K_t^{S'}(\theta,\cdot)
    = \mathcal{N}\!\Bigl(\theta-h_t\nabla L_{S'}(\theta),\ \tfrac{2h_t}{\beta_t}I_d\Bigr),
    \qquad t=0,\dots,T-1,
\]
are measurable. For fixed $\theta$ and $t$, the kernels $K_t^{S}(\theta,\cdot)$ and
$K_t^{S_I}(\theta,\cdot)$ are Gaussians with the same covariance
$\tfrac{2h_t}{\beta_t}I_d$ and means that differ by
$h_t\bigl(\nabla L_{S_I}(\theta)-\nabla L_S(\theta)\bigr)$. The KL divergence
between two Gaussians with a common covariance $\Sigma$ is
$\tfrac12(\Delta m)^\top\Sigma^{-1}\Delta m$, so
\[
    D_{\mathrm{KL}}\bigl(K_t^{S}(\theta,\cdot)\,\|\,K_t^{S_I}(\theta,\cdot)\bigr)
    = \frac12\cdot\frac{\beta_t}{2h_t}\cdot h_t^2
      \bigl\|\nabla L_{S_I}(\theta)-\nabla L_S(\theta)\bigr\|^2
    = \frac{\beta_t h_t}{4}\bigl\|\nabla L_{S_I}(\theta)-\nabla L_S(\theta)\bigr\|^2 .
\]

Both chains are Markov with the same initial law $\mu_0$ and use the same
Gaussian noise structure, so the chain rule for KL divergence on Markov paths gives
\[
    D_{\mathrm{KL}}\bigl(\mathbb{P}^{S}\,\|\,\mathbb{P}^{S_I}\bigr)
    = \underbrace{D_{\mathrm{KL}}(\mu_0\|\mu_0)}_{=0}
    + \sum_{t=0}^{T-1}\mathbb{E}_{\theta_t\sim\mu_t^{S}}
      \Bigl[D_{\mathrm{KL}}\bigl(K_t^{S}(\theta_t,\cdot)\,\|\,K_t^{S_I}(\theta_t,\cdot)\bigr)\Bigr].
\]
The outer expectation is under the marginal $\mu_t^{S}$ of the first path law,
which is why $G_T$ involves only the full-data chain. Thus:
\[
    D_{\mathrm{KL}}\bigl(\mathbb{P}^{S}\,\|\,\mathbb{P}^{S_I}\bigr) = G_T(I).
\]

By Jensen's inequality, for every $\theta$,
$\|\nabla L_{S'}(\theta)\|^2\le \tfrac{1}{|S'|}\sum_{z\in S'}\|\nabla\ell(\theta,z)\|^2$,
so
\[
    \bigl\|\nabla L_{S_I}(\theta)-\nabla L_S(\theta)\bigr\|^2
    \le 2\|\nabla L_{S_I}(\theta)\|^2+2\|\nabla L_S(\theta)\|^2
    \le \frac{2}{m}\sum_{i\in I}\|\nabla\ell(\theta,Z_i)\|^2
      +\frac{2}{n}\sum_{i=1}^{n}\|\nabla\ell(\theta,Z_i)\|^2 .
\]
The integrability assumption on $\nabla\ell(\theta,Z_i)$ under $\mu_t^{S}$, for
$t\le T$, therefore ensures that every term of $G_T(I)$ is finite.

The map $(\theta_0,\dots,\theta_T)\mapsto\theta_T$ pushes $\mathbb{P}^{S}$ and
$\mathbb{P}^{S_I}$ forward to $\mu_T^{S}$ and $\mu_T^{S_I}$. Data processing for
KL divergence then yields
\[
    D_{\mathrm{KL}}\bigl(\Alg_n(S)\,\|\,\Alg_m(S_I)\bigr)
    = D_{\mathrm{KL}}\bigl(\mu_T^{S}\,\|\,\mu_T^{S_I}\bigr)
    \le D_{\mathrm{KL}}\bigl(\mathbb{P}^{S}\,\|\,\mathbb{P}^{S_I}\bigr)
    = G_T(I). \qedhere
\]
\end{proof}

\section{Proof of the quadratic program formulation of subset selection by optimizing the gradient trajectory}\label{app:qp_formulation}
\begin{proof}
Write $g_i(\theta):=\nabla\ell(\theta,Z_i)$, $\bar g(\theta):=\nabla L_S(\theta)=\frac1n\sum_{i=1}^n g_i(\theta)$,
so that $\zeta_i=g_i-\bar g$, and $w_t:=\beta_th_t/4$. Since $\sum_{j=1}^n\zeta_j(\theta)=0$
for every $\theta$,
\[
  (C\mathbf{1})_i=\sum_{t=0}^{T-1}w_t\,\mathbb{E}_{\theta\sim\mu_t^S}
  \Bigl\langle\zeta_i(\theta),\sum_{j=1}^n\zeta_j(\theta)\Bigr\rangle=0 .
\]
For $v\in\mathbb{R}^n$, linearity gives
\[
  v^\top Cv=\sum_{t=0}^{T-1}w_t\,\mathbb{E}_{\theta\sim\mu_t^S}
  \Bigl\|\sum_{i=1}^n v_i\,\zeta_i(\theta)\Bigr\|^2\ge0,
\]
so $C\succeq0$. Finally, for $|I|=m$,
\[
  \nabla L_{S_I}(\theta)-\nabla L_S(\theta)
  =\frac1m\sum_{i\in I}g_i(\theta)-\bar g(\theta)
  =\frac1m\sum_{i\in I}\zeta_i(\theta),
\]
and therefore
\[
  G_T(I)=\sum_{t=0}^{T-1}w_t\,\mathbb{E}_{\theta\sim\mu_t^S}
  \Bigl\|\frac1m\sum_{i\in I}\zeta_i(\theta)\Bigr\|^2
  =\frac1{m^2}\sum_{i,j\in I}C_{ij}
  =\frac1{m^2}\mathbf{1}_I^\top C\mathbf{1}_I .
\]
Since $I\mapsto\mathbf{1}_I$ is a bijection from subsets of size $m$ onto
$\mathcal{B}_m$, minimizing $G_T$ over such subsets is the stated integer program.
\end{proof}

\section{Construction and analysis of the Swap algorithm}\label{app:simple-swap-gap}
\paragraph{Setup.}
Write $w_t:=\beta_th_t/4$. For $\alpha\in\mathcal{B}_m$, let
$D(\alpha)=\frac1{m^2}\alpha^\top C\alpha$ be the objective of
\Cref{thm:qp_formulation}, so that $D(\mathbf 1_I)=G_T(I)$, and let
\[
    s(\alpha):=\nabla D(\alpha)=\frac{2}{m^{2}}\,C\alpha,
    \qquad D(\alpha)=\tfrac12\,\alpha^\top s(\alpha).
\]
Since $C\succeq0$ and $C\mathbf 1=0$, $D$ is convex and already centered. Fix
\[
    \alpha^\star\in\argmin_{\alpha\in\mathcal{B}_m}D(\alpha),
    \qquad D^\star:=D(\alpha^\star),
    \qquad r:=\min\{m,n-m\},
\]
and write $[x]_+:=\max\{x,0\}$. Given an iterate $\alpha_k=\mathbf 1_{I_k}$ with
stored $s_k:=s(\alpha_k)$, let
\[
    i_k\in\argmax_{i\in I_k}s_k(i),
    \qquad
    j_k\in\argmin_{j\notin I_k}s_k(j),
    \qquad
    \gamma_k:=s_k(i_k)-s_k(j_k),
    \qquad
    \delta_k:=e_{j_k}-e_{i_k}.
\]
The pair $(i_k,j_k)$ minimizes the linearized change $s_k(j)-s_k(i)$ over all
exchanges.

\begin{algorithm}[H]
\caption{Gradient-ranked single-swap descent (indicator form)}\label{alg:swap-appendix}
\begin{algorithmic}[1]
\State Initialize $\alpha_0:=\mathbf 1_{I_0}$ with $|I_0|=m$, and query $C\alpha_0$ to store $s_0$.
\State Compute $D(\alpha_0)$ from $C\alpha_0$ and set $\underline D_{-1}:=-\infty$.
\For{$k=0,1,\ldots$}
    \State Choose $i_k$, $j_k$, $\gamma_k$, and $\delta_k$ as above.
    \State Set $\underline D_k:=\max\{\underline D_{k-1},D(\alpha_k)-r[\gamma_k]_+\}$ and
    $\widehat H_k:=D(\alpha_k)-\underline D_k$.
    \If{$\gamma_k\le0$} \State \Return $(\alpha_k,\widehat H_k)$. \EndIf
    \State Set $\widetilde\alpha_k:=\alpha_k+\delta_k$, query $C\widetilde\alpha_k$, and store
    $\widetilde s_k:=s(\widetilde\alpha_k)$.
    \State Compute $\Delta_k:=\frac12\delta_k^{\top}(s_k+\widetilde s_k)$.
    \If{$\Delta_k<0$}
        \State Set $(\alpha_{k+1},s_{k+1}):=(\widetilde\alpha_k,\widetilde s_k)$ and store
        $D(\alpha_{k+1}):=D(\alpha_k)+\Delta_k$.
    \Else
        \State \Return $(\alpha_k,\widehat H_k)$.
    \EndIf
\EndFor
\end{algorithmic}
\end{algorithm}

This is \Cref{alg:gradient-ranked-swap} written on indicator vectors, with an early
exit and the bookkeeping of $\underline D_k$ and $\widehat H_k$. Neither changes the
iterates, as shown below.

\subsection{Proof of the swap gap}
\begin{proof}
Let $I'=(I_k\setminus\{i_k\})\cup\{j_k\}$, so that $\widetilde\alpha_k=\mathbf 1_{I'}$.
Since $D$ is quadratic,
\[
  \Delta_k=\tfrac12\delta_k^\top(s_k+\widetilde s_k)
  =\delta_k^\top s_k+\frac1{m^2}\delta_k^\top C\delta_k
  =D(\widetilde\alpha_k)-D(\alpha_k)=G_T(I')-G_T(I_k),
\]
and $\delta_k^\top s_k=-\gamma_k$ with $\delta_k^\top C\delta_k\ge0$, so
$\Delta_k\ge-\gamma_k$. Hence a candidate is accepted if and only if
$G_T(I')<G_T(I_k)$. Moreover, if $\gamma_k\le0$ then $\Delta_k\ge0$, so the
candidate would be rejected anyway. The early exit therefore does not change the
iterates, and the algorithm coincides with \Cref{alg:gradient-ranked-swap}.

(i) Since $1\le m\le n-1$, both $I_k$ and its complement are nonempty, so
$i_k\in I_k$ and $j_k\notin I_k$ exist for every tie-breaking rule. Hence
$|I'|=m$, and by induction every $I_k$ has size $m$. Every accepted swap satisfies
$G_T(I_{k+1})<G_T(I_k)$, so the values $G_T(I_0)>G_T(I_1)>\cdots$ are strictly
decreasing and the accepted subsets are pairwise distinct. There are only
$\binom nm$ subsets of size $m$, so at most $\binom nm-1$ swaps are accepted. The
first rejected candidate (or the early exit) then makes the algorithm return
$I_{\mathrm{out}}=I_k$.

(ii) By (i), $G_T(I_{\mathrm{out}})\le G_T(I_0)$ for every $I_0$ and every
tie-breaking rule, so $\mathbb{E}[G_T(I_{\mathrm{out}})]\le\mathbb{E}[G_T(I_0)]$.
For $I_0$ uniform on the subsets of size $m$, we have $\mathbb{P}(i\in I_0)=\frac mn$
and $\mathbb{P}(i,j\in I_0)=\frac{m(m-1)}{n(n-1)}$ for $i\ne j$. Therefore
\[
  \mathbb{E}[G_T(I_0)]
  =\frac1{m^2}\Bigl[\frac mn\operatorname{tr}C
  +\frac{m(m-1)}{n(n-1)}\sum_{i\ne j}C_{ij}\Bigr].
\]
Since $C\mathbf{1}=0$, we have $\sum_{i\ne j}C_{ij}=\mathbf{1}^\top C\mathbf{1}-\operatorname{tr}C=-\operatorname{tr}C$, and hence
\[
  \mathbb{E}[G_T(I_0)]
  =\frac{\operatorname{tr}C}{m^2}\cdot\frac mn\Bigl(1-\frac{m-1}{n-1}\Bigr)
  =\frac{n-m}{m\,n\,(n-1)}\operatorname{tr}C .
\]
Finally, with $\zeta_i=\nabla\ell(\cdot,Z_i)-\nabla L_S$,
\[
  \operatorname{tr}C=\sum_{i=1}^nC_{ii}
  =\sum_{t=0}^{T-1}w_t\sum_{i=1}^n\mathbb{E}_{\theta\sim\mu_t^S}\|\zeta_i(\theta)\|^2
  =n\sum_{t=0}^{T-1}w_tV_t,
\]
so that $\mathbb{E}[G_T(I_{\mathrm{out}})]\le\mathbb{E}[G_T(I_0)]=\frac{n-m}{m(n-1)}\sum_{t=0}^{T-1}w_tV_t$.
\end{proof}

\section{Experimental details}\label{app:experimental_details}
\subsection{Caveat on the train / validation split}\label{app:caveat_train_val}
The reference trajectory is trained on a single random $90\%/10\%$ split (seed 42),
whereas the selection experiments use five seeds, each with its own split. Only one
seed therefore shares its pool with the reference run. For the other four, the
selection pool $S_s$ and the reference pool $S_{\mathrm{ref}}$ are different
$45{,}000$-example draws from the same $50{,}000$ images, overlapping on about $90\%$
of their elements. For these seeds, both the checkpoints used to compute
$\widehat{G}_T$ and the full-data model $\Alg(S)$ in the faithfulness gap of
Figure~\ref{fig:corr_d_alpha} come from $S_{\mathrm{ref}}$ rather than $S_s$. We did
not retrain a reference trajectory per seed because of the cost ($5\times$ the most
expensive step).
\subsection{More empirical details}

For each (dataset, architecture) pair we train one full-data model on the pool of seed~$42$
with the protocol of \cref{sec:experiments}, and store the parameters at the end of selected
epochs, together with the step index and learning rate $\eta_k$ of that epoch.
The \emph{same} reference trajectory is used for all five seeds (see caveat in \cref{app:caveat_train_val}).

We use the SGD objective: $w_k=\eta_k/\sum_j\eta_j$, where $\eta_k$ is the cosine-schedule
learning rate of the epoch $e_k$ in which checkpoint $k$ was saved,
$\eta_k=\tfrac12\eta_0\bigl(1+\cos(\pi(e_k-1)/300)\bigr)$ with $\eta_0=0.01$.
The weights are normalized to sum to one, so $\widehat G_T$ is a weighted mean
over checkpoints. This rescales $\widehat G_T$ but changes no swap decision.

\paragraph{Checkpoints used for selection.}
For $\tau$ checkpoints we take the stored epochs closest to $\tau$ points equally spaced in
optimizer-step time between epoch~$1$ and epoch~$300$. Hence $\tau=1$ gives the epoch-$1$
checkpoint, and $\tau=4$ gives approximately epochs $1,100,200,300$. Baseline comparison experiments (\cref{fig:baseline_comparison})
use $\tau=1$; \cref{app:checkpoint_ablation} varies $\tau$.

\section{Empirical correlation between faithfulness error and trajectory gradient discrepancy}\label{app:d_alpha_corr}

Below, we provide results for other settings showing the correlation between the faithfulness error and the estimate of the proxy, $\sqrt{\widehat{G}_T(I)}$. We use similar settings as those explained in \cref{sec:experiments} and \cref{app:experimental_details}. The estimate of $G_T$ is computed using a single training run on $45,000$ sampled points from the training dataset.

\paragraph{Reference trajectory.} For each of the six (dataset, architecture) pairs, we train one model on the 45,000 sampled training points using SGD, batch size $128$, initial learning rate $0.01$ annealed by a cosine schedule over $300$ epochs, and we store the parameters $\theta_k$. This run plays the role of $\Alg(S)$ and is the only training on $S$ that the proxy needs.

\paragraph{Estimating the discrepancy.} For a candidate subset $I$ we evaluate
\[
\widehat{G}_T(I) \;=\; \sum_{k} w_k \,\bigl\lVert \nabla L_{S_I}(\theta_k) - \nabla L_{S}(\theta_k) \bigr\rVert_2^2 ,
\]
the sum running over the stored checkpoints, with weights $w_k \propto \eta_k$ normalised to sum to one. The two mean gradients are computed exactly, over \emph{all} trainable parameters and over the whole pool. Examples are viewed through the deterministic test transform and the network is in evaluation mode, so each per-example gradient is a well-defined function of $\theta_k$ alone. This replaces the expectation over $\mu^S_t$ by its value at a single realised trajectory.

\paragraph{Candidate subsets and faithfulness error.} The points of each panel are the subsets returned by the selection methods of \cref{sec:experiments}, including swap-descent variant, at retention fraction $0.1$ and over the $5$ seeds; one point is one (method, seed) pair, with no averaging. Every subset is retrained from a shared initialization with the same optimizer, schedule and number of epochs as the reference run, and the official test set is evaluated once, after the final epoch. 
Each panel fixes the dataset, the architecture \emph{and} the selection ratio is set to $10\%$.

\begin{figure}[htbp]
    \centering
    \begin{subfigure}[b]{0.32\textwidth}
        \centering
        \includegraphics[width=\textwidth]{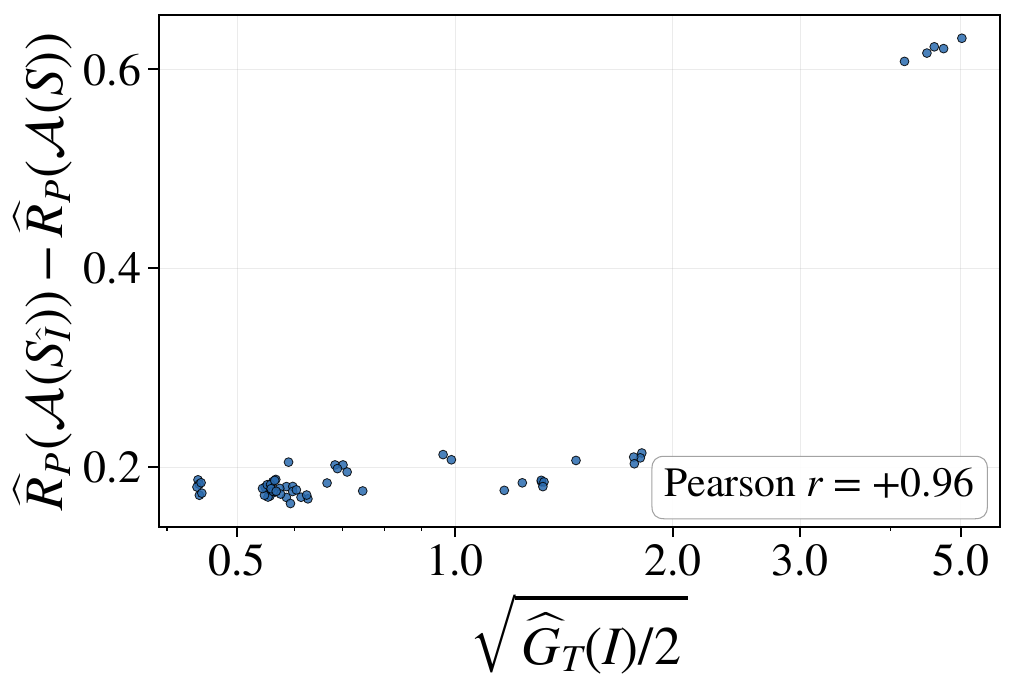}
        \caption{VGG16}
        \label{fig:sub1}
    \end{subfigure}%
    \begin{subfigure}[b]{0.32\textwidth}
        \centering
        \includegraphics[width=\textwidth]{figures/cifar10_resnet18_f0.1_riskgap_vs_sqrtG.pdf}
        \caption{ResNet18}
        \label{fig:sub2}
    \end{subfigure}
    \hfill
    \begin{subfigure}[b]{0.32\textwidth}
        \centering
        \includegraphics[width=\textwidth]{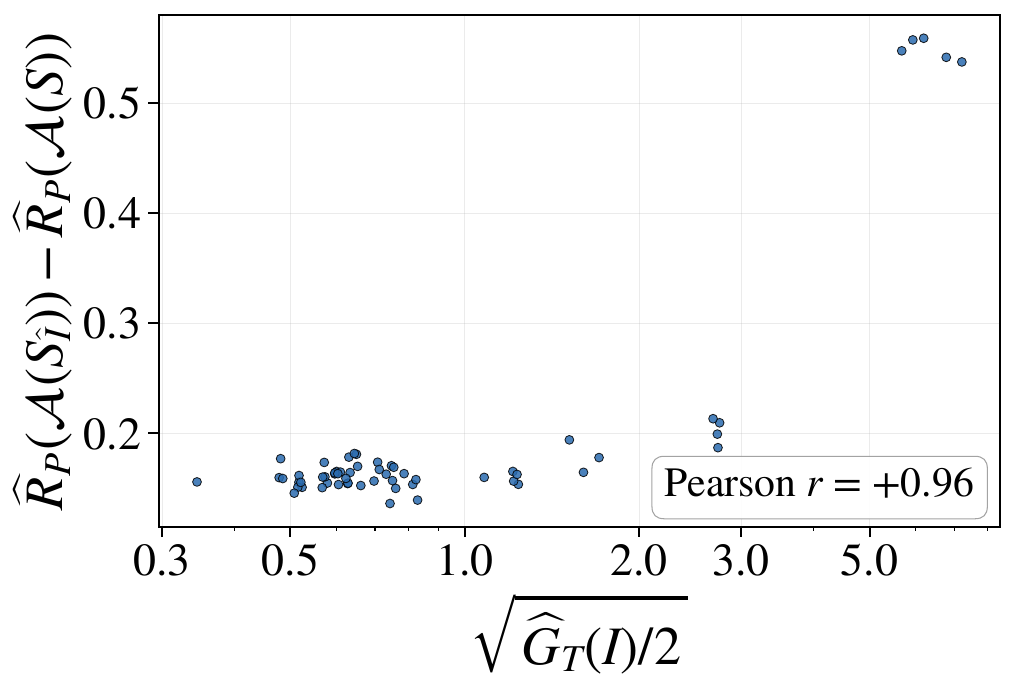}
        \caption{Inception}
        \label{fig:sub3}
    \end{subfigure}
    
    \vspace{0.3cm}
    \begin{subfigure}[b]{0.32\textwidth}
        \centering
        \includegraphics[width=\textwidth]{figures/cifar100_vgg16_f0.1_riskgap_vs_sqrtG.pdf}
        \caption{VGG16}
        \label{fig:sub4}
    \end{subfigure}
    \hfill
    \begin{subfigure}[b]{0.32\textwidth}
        \centering
        \includegraphics[width=\textwidth]{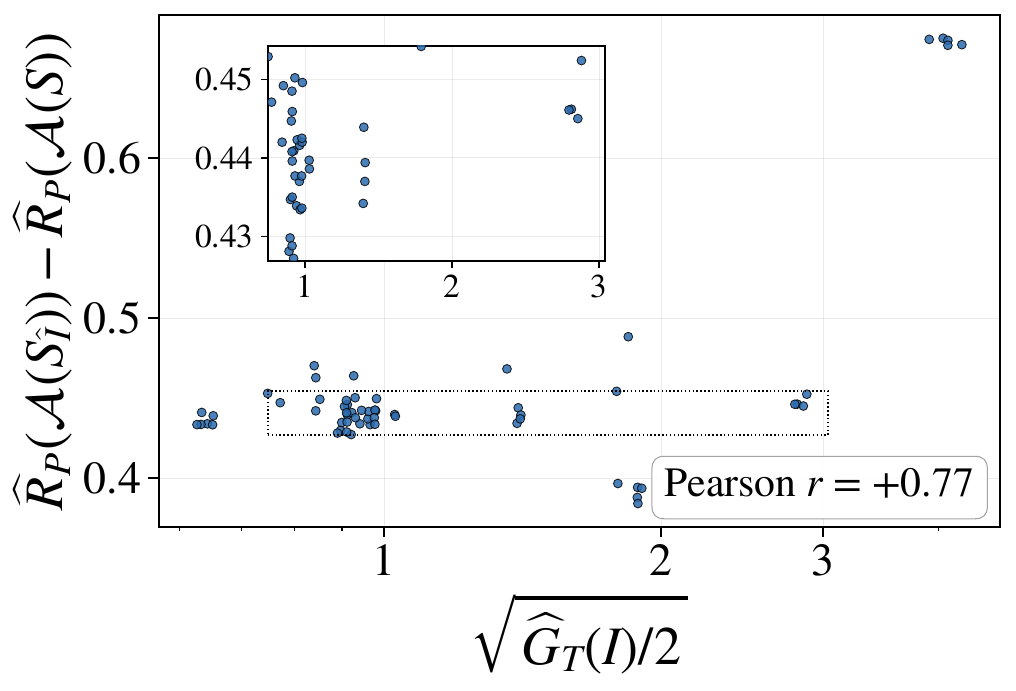}
        \caption{ResNet18}
        \label{fig:sub5}
    \end{subfigure}
    \hfill
    \begin{subfigure}[b]{0.32\textwidth}
        \centering
        \includegraphics[width=\textwidth]{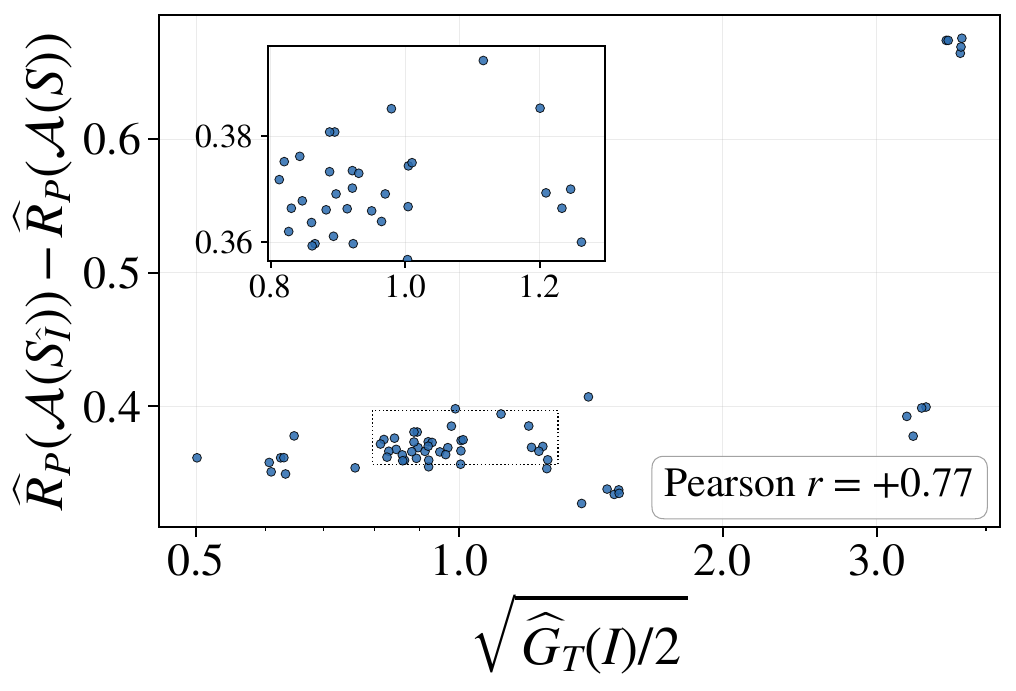}
        \caption{Inception}
        \label{fig:sub6}
    \end{subfigure}

    \caption{Correlation between the empirical faithfulness gap and the estimate of the proxy $\sqrt{\widehat{G}_T(I)/2}$. Consistent correlation is shown across datasets (CIFAR10, first row, CIFAR100 second row) and architecture. The zoomed insets show the boxed region.}
    \label{fig:corr_g_faithful} 
\end{figure}

\paragraph{Results.} \cref{fig:corr_g_faithful} reports the Pearson correlation between the estimated faithfulness and the estimated $\widehat{G}_T$, pooled over methods and seeds within each panel. It exceeds $0.70$ in all six settings, so across architectures and datasets the subsets with the largest trajectory gradient discrepancy are also the ones whose retrained model departs most from full-data training.

\section{Gradient-Swap with warm starts}\label{app:warm_start}

In order to show the benefit of minimizing the proxy $\widehat{G}_T$ as a way to improve the faithfulness of a subset, we ask whether, starting from a subset obtained by a given data selection method---we pick GraphCut (one of the best-performing baselines), Random, Herding, and EL2N (one of the least-performing ones)---applying swaps of Gradient Swap improves both the $\widehat{G}_T$ objective and downstream accuracy.
As expected from \cref{thm:simple-swap-gap}, the swaps lower $\widehat{G}_T$ for every starting selector, even though $\tau{=}10$ checkpoints are used for the swaps and all $300$ are used for evaluation.
The downstream accuracy results reflect the one-sided nature of Corollary~4.2: once a selector already lies inside the low-$\widehat{G}_T$ faithfulness envelope (GraphCut, Herding, and Random), the upper bound is already small and further reducing $\widehat{G}_T$ leaves test accuracy unchanged within one standard deviation.
Conversely, for EL2N, whose subsets severely violate trajectory fidelity ($\widehat{G}_T = 81.22$), just $K{=}500$ swaps reduce $\widehat{G}_T$ by $16\%$ and improve test accuracy by $2.6$ percentage points, although it remains far from the other selectors.
This confirms that $\widehat{G}_T$ acts as a one-sided safety certificate: it is most informative for diagnosing and repairing unfaithful subsets, whereas further decreases in the low-discrepancy regime do not translate into additional accuracy gains.
\begin{table}
\centering
\caption{Warm start of full-network Gradient Swap on cifar10 / resnet18 at 10\% of the pool. Accuracy is the last logged test epoch. $G_T$ is the full-network objective on every reference checkpoint. The after columns are exact full-network Gradient Swap at $\tau=10$ started from the baseline subset. Mean and standard deviation are over $5$ seeds.}\label{tab:warm_start}
\begin{tabular}{lcccc}
\toprule
Method & Acc. before & Acc. after & $G_T$ before & $G_T$ after \\
\midrule
GraphCut & 72.60 $\pm$ 0.54 & 72.16 $\pm$ 0.60 & 3.414 $\pm$ 0.2469 & 3.13 $\pm$ 0.2379 \\
Herding & 71.74 $\pm$ 1.56 & 71.56 $\pm$ 0.84 & 0.8044 $\pm$ 0.1356 & 0.7549 $\pm$ 0.1253 \\
Random & 71.66 $\pm$ 0.69 & 72.07 $\pm$ 0.11 & 0.7889 $\pm$ 0.2079 & 0.6516 $\pm$ 0.1152 \\
EL2N & 29.66 $\pm$ 1.12 & \textbf{32.29 $\pm$ 1.11} & 81.22 $\pm$ 4.636 & \textbf{68.61 $\pm$ 3.138} \\
\bottomrule
\end{tabular}
\end{table}

\section{Ablation study on the number of used checkpoints}\label{app:checkpoint_ablation}
\cref{tab:ablation_ckpt_cifar10,tab:ablation_ckpt_cifar100} show the results of the checkpoint ablation on CIFAR10 and CIFAR100. Using the same setting as the aforementioned experiments, we can see that the number of used checkpoints does not consistently affect test accuracy for three architectures, motivating the use of $\tau=1$ as a computational simplification for the estimation of $G_T$.

\begin{table}
\centering
\caption{Test accuracy (\%) of full-network Gradient Swap on CIFAR10. Values are mean $\pm$ standard deviation over seeds.}\label{tab:ablation_ckpt_cifar10}
\begin{tabular}{llcccc}
\toprule
Architecture & $\tau$ & \multicolumn{4}{c}{Test accuracy} \\
\cmidrule(lr){3-6}
 & & 10\% & 20\% & 40\% & 50\% \\
\midrule
ResNet-18 & 1 & 69.57 $\pm$ 1.00 & 80.12 $\pm$ 0.80 & 87.02 $\pm$ 0.25 & 88.86 $\pm$ 0.30 \\
ResNet-18 & 4 & 69.77 $\pm$ 0.85 & 80.37 $\pm$ 0.51 & 87.05 $\pm$ 0.39 & 88.91 $\pm$ 0.18 \\
ResNet-18 & 10 & 69.81 $\pm$ 1.48 & 80.13 $\pm$ 0.77 & 87.11 $\pm$ 0.34 & 88.69 $\pm$ 0.31 \\
ResNet-18 & 15 & 69.06 $\pm$ 1.35 & 80.30 $\pm$ 0.80 & 87.05 $\pm$ 0.23 & 88.64 $\pm$ 0.23 \\
ResNet-18 & 20 & 69.86 $\pm$ 0.72 & 80.34 $\pm$ 0.65 & 86.97 $\pm$ 0.21 & 88.73 $\pm$ 0.22 \\
\midrule
VGG-16 & 1 & 73.34 $\pm$ 0.71 & 80.86 $\pm$ 0.55 & 86.31 $\pm$ 0.26 & 87.92 $\pm$ 0.22 \\
VGG-16 & 4 & 72.74 $\pm$ 0.78 & 80.57 $\pm$ 0.47 & 86.34 $\pm$ 0.36 & 87.68 $\pm$ 0.23 \\
VGG-16 & 10 & 73.41 $\pm$ 0.53 & 80.63 $\pm$ 0.25 & 86.19 $\pm$ 0.13 & 87.68 $\pm$ 0.25 \\
VGG-16 & 15 & 72.84 $\pm$ 0.85 & 80.72 $\pm$ 0.58 & 86.22 $\pm$ 0.29 & 87.91 $\pm$ 0.23 \\
VGG-16 & 20 & 72.40 $\pm$ 0.94 & 80.74 $\pm$ 0.41 & 86.39 $\pm$ 0.44 & 87.79 $\pm$ 0.28 \\
\midrule
Inception & 1 & 75.34 $\pm$ 1.11 & 83.55 $\pm$ 0.31 & 88.43 $\pm$ 0.21 & 89.46 $\pm$ 0.24 \\
Inception & 4 & 75.22 $\pm$ 1.13 & 83.71 $\pm$ 0.23 & 88.10 $\pm$ 0.57 & 89.33 $\pm$ 0.34 \\
Inception & 10 & 75.35 $\pm$ 0.78 & 83.75 $\pm$ 0.30 & 88.27 $\pm$ 0.29 & 89.43 $\pm$ 0.22 \\
Inception & 15 & 75.00 $\pm$ 0.80 & 83.62 $\pm$ 0.44 & 88.47 $\pm$ 0.23 & 89.61 $\pm$ 0.29 \\
Inception & 20 & 75.93 $\pm$ 0.49 & 83.61 $\pm$ 0.25 & 88.44 $\pm$ 0.31 & 89.39 $\pm$ 0.35 \\
\bottomrule
\end{tabular}
\end{table}

\begin{table}
\centering
\caption{Test accuracy (\%) of full-network Gradient Swap on CIFAR100. Values are mean $\pm$ standard deviation over seeds.}\label{tab:ablation_ckpt_cifar100}
\begin{tabular}{llcccc}
\toprule
Architecture & $\tau$ & \multicolumn{4}{c}{Test accuracy} \\
\cmidrule(lr){3-6}
 & & 10\% & 20\% & 40\% & 50\% \\
\midrule
ResNet-18 & 1 & 28.79 $\pm$ 0.76 & 42.31 $\pm$ 0.54 & 57.32 $\pm$ 0.76 & 61.79 $\pm$ 0.39 \\
ResNet-18 & 4 & 28.71 $\pm$ 0.64 & 42.70 $\pm$ 0.55 & 57.05 $\pm$ 0.32 & 61.77 $\pm$ 0.33 \\
ResNet-18 & 10 & 28.66 $\pm$ 0.44 & 42.43 $\pm$ 0.76 & 57.18 $\pm$ 0.72 & 61.81 $\pm$ 0.28 \\
ResNet-18 & 15 & 28.92 $\pm$ 0.75 & 42.49 $\pm$ 0.44 & 57.27 $\pm$ 0.80 & 61.62 $\pm$ 0.76 \\
ResNet-18 & 20 & 28.79 $\pm$ 0.89 & 42.58 $\pm$ 0.48 & 57.33 $\pm$ 0.47 & 61.92 $\pm$ 0.42 \\
\midrule
VGG-16 & 1 & 30.51 $\pm$ 0.20 & 44.04 $\pm$ 0.55 & 56.60 $\pm$ 0.52 & 60.01 $\pm$ 0.17 \\
VGG-16 & 4 & 30.25 $\pm$ 0.93 & 43.72 $\pm$ 0.74 & 56.37 $\pm$ 0.34 & 60.02 $\pm$ 0.37 \\
VGG-16 & 10 & 30.74 $\pm$ 0.83 & 43.91 $\pm$ 0.51 & 56.71 $\pm$ 0.46 & 59.75 $\pm$ 0.40 \\
VGG-16 & 15 & 30.73 $\pm$ 0.65 & 43.99 $\pm$ 0.33 & 56.67 $\pm$ 0.56 & 59.81 $\pm$ 0.48 \\
VGG-16 & 20 & 30.67 $\pm$ 0.38 & 44.04 $\pm$ 0.62 & 56.65 $\pm$ 0.26 & 59.70 $\pm$ 0.13 \\
\midrule
Inception & 1 & 36.97 $\pm$ 0.70 & 51.58 $\pm$ 0.44 & 62.69 $\pm$ 0.43 & 65.69 $\pm$ 0.58 \\
Inception & 4 & 36.78 $\pm$ 1.01 & 51.73 $\pm$ 0.92 & 62.88 $\pm$ 0.32 & 65.73 $\pm$ 0.46 \\
Inception & 10 & 36.50 $\pm$ 0.91 & 52.02 $\pm$ 0.78 & 62.65 $\pm$ 0.56 & 65.92 $\pm$ 0.37 \\
Inception & 15 & 36.53 $\pm$ 1.02 & 51.84 $\pm$ 0.21 & 62.47 $\pm$ 0.48 & 66.00 $\pm$ 0.47 \\
Inception & 20 & 36.71 $\pm$ 0.83 & 51.71 $\pm$ 0.88 & 62.44 $\pm$ 0.30 & --- \\
\bottomrule
\end{tabular}
\end{table}

\section{Johnson-Lindenstrauss Embeddings Ablation}\label{app:jl_embeddings_ablation}

Following \citet{xia2024less}, we attempt dimensionality reduction techniques (Johnson Lindenstrauss Embeddings \citep{johnson1984extensions}) on the gradient to reduce the computational and storage cost of \cref{alg:gradient-ranked-swap}. 
Full-network Gradient Swap stores one aggregate gradient per checkpoint. A count sketch of width \(p\) replaces each aggregate by its image in \(\mathbb{R}^p\), and the swap test is exact for that sketched objective. In \cref{tab:jl-tau1,tab:jl-tau4}, we compare \(p \in \{1024, 2048, 4096\}\) with the exact aggregates $p=d$ on CIFAR-10 with ResNet-18 ($d \approx 10^7$), at checkpoint counts \(\tau \in \{1, 4\}\) and subset fractions \(10\%\), \(20\%\), \(40\%\), and \(50\%\). Each entry is the mean test accuracy and the standard deviation over five seeds.
The sketch does not change the selected subset's accuracy. At every fraction and both values of \(\tau\), the three sketched runs sit on the exact run: the largest gap is \(0.60\) points (\(\tau = 1\), \(p = 2048\), \(10\%\)), and that gap is smaller than the seed standard deviation of either run. From \(40\%\) upward the gaps are about \(0.1\) points. Width \(1024\) is already enough. Increasing \(p\) to \(2048\) or \(4096\) does not produce a further gain, and the same match holds when the trajectory grows from one checkpoint to four. The embedding can therefore be used in place of the exact aggregates.

\begin{table}[htbp]
  \centering
  \caption{Test accuracy (\%) on CIFAR-10 with ResNet-18 at \(\tau = 1\). \(p = d\) is the exact aggregate. Mean and standard deviation over five seeds.}
  \label{tab:jl-tau1}
  \begin{tabular}{lcccc}
    \toprule
    Fraction & $p=d$ & \(p = 1024\) & \(p = 2048\) & \(p = 4096\) \\
    \midrule
    \(10\%\) & \(69.57 \pm 1.00\) & \(69.50 \pm 1.37\) & \(68.97 \pm 0.76\) & \(69.66 \pm 0.85\) \\
    \midrule
    \(20\%\) & \(80.12 \pm 0.80\) & \(80.44 \pm 0.54\) & \(80.39 \pm 0.55\) & \(79.94 \pm 0.37\) \\
    \midrule
    \(40\%\) & \(87.02 \pm 0.25\) & \(87.02 \pm 0.12\) & \(86.96 \pm 0.53\) & \(87.12 \pm 0.16\) \\
    \midrule
    \(50\%\) & \(88.86 \pm 0.30\) & \(88.74 \pm 0.16\) & \(88.93 \pm 0.29\) & \(88.75 \pm 0.17\) \\
    \bottomrule
  \end{tabular}
\end{table}

\begin{table}[htbp]
  \centering
  \caption{Test accuracy (\%) on CIFAR-10 with ResNet-18 at \(\tau = 4\). \(p = d\) is the exact aggregate. Mean and standard deviation over five seeds.}
  \label{tab:jl-tau4}
  \begin{tabular}{lcccc}
    \toprule
    Fraction & \(p = d\) & \(p = 1024\) & \(p = 2048\) & \(p = 4096\) \\
    \midrule
    \(10\%\) & \(69.77 \pm 0.85\) & \(69.79 \pm 0.39\) & \(70.04 \pm 0.70\) & \(69.76 \pm 0.65\) \\
    \midrule
    \(20\%\) & \(80.37 \pm 0.51\) & \(79.93 \pm 0.73\) & \(80.28 \pm 0.47\) & \(80.04 \pm 0.51\) \\
    \midrule
    \(40\%\) & \(87.05 \pm 0.39\) & \(87.10 \pm 0.22\) & \(87.08 \pm 0.32\) & \(87.09 \pm 0.15\) \\
    \midrule
    \(50\%\) & \(88.91 \pm 0.18\) & \(88.86 \pm 0.02\) & \(88.84 \pm 0.16\) & \(88.96 \pm 0.14\) \\
    \bottomrule
  \end{tabular}
\end{table}

\section{Selected examples for each selection method}
For each dataset and architecture we show the training images selected at 10\% of the pool (seed 42). Each row is one method: Gradient Swap using full-network gradients and a single checkpoint $(\tau=1)$, then Random, Herding, Moderate, EL2N, MoSo, GraphCut, and Facility Location. From that method's saved subset we draw six examples without replacement. The stored indices are positions in the 90\% training pool; we map them back to the original training set by replaying the same seed-42 split, and we display the raw images with no crop, flip, or normalization. The label under each image is its class.

\includegraphics[width=0.95\textwidth]{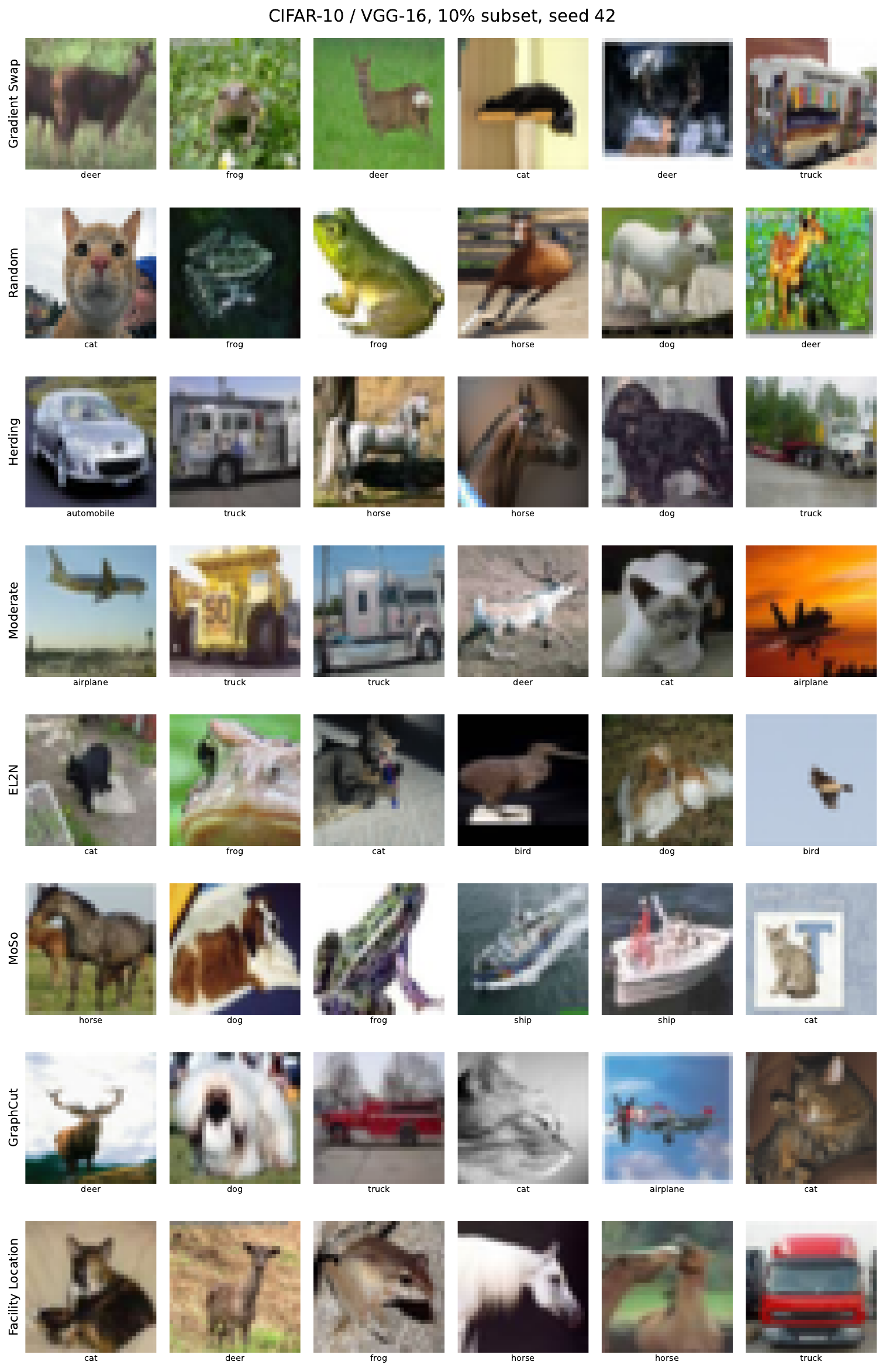}
\includegraphics[width=0.95\textwidth]{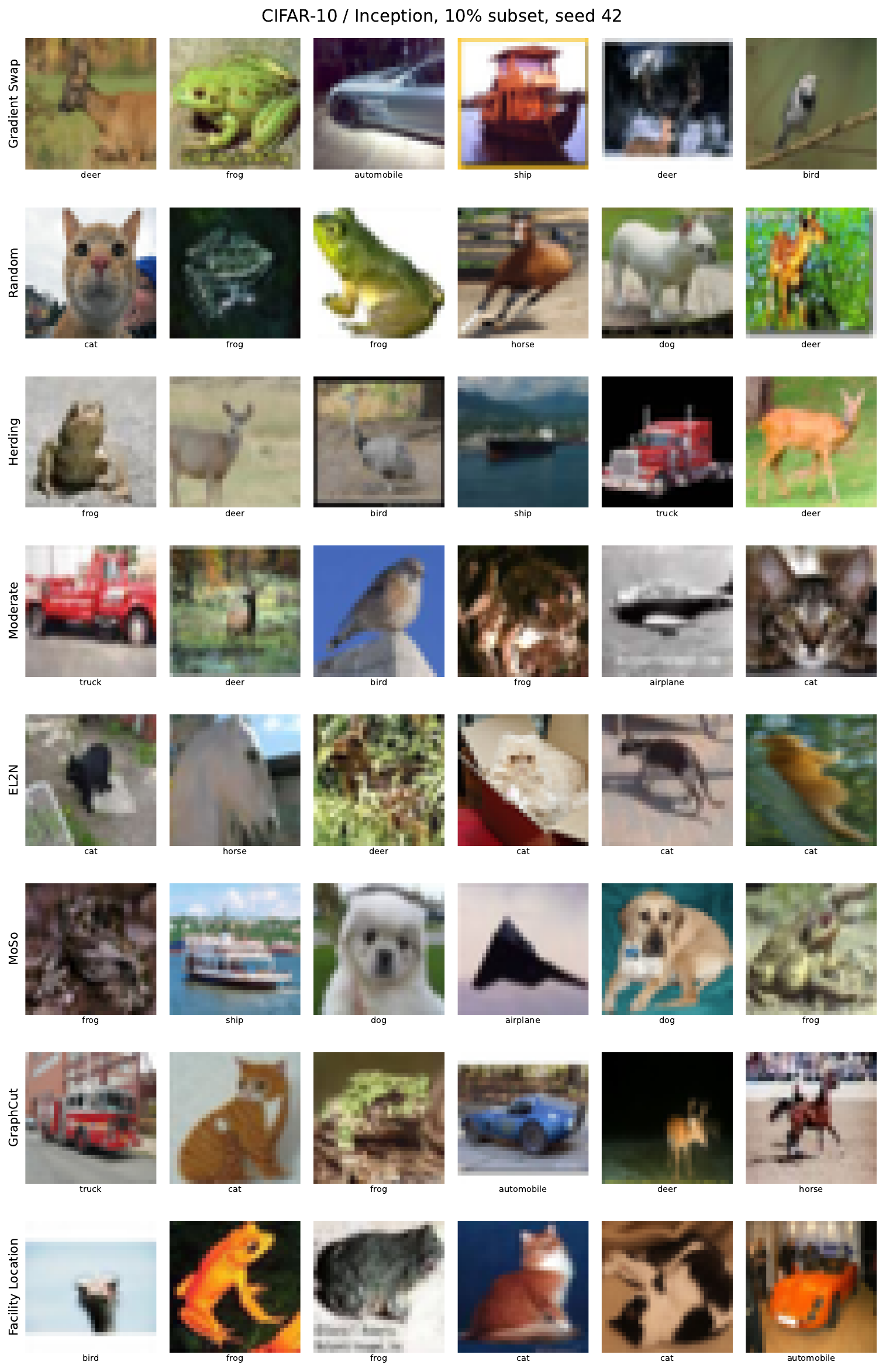}
\includegraphics[width=0.95\textwidth]{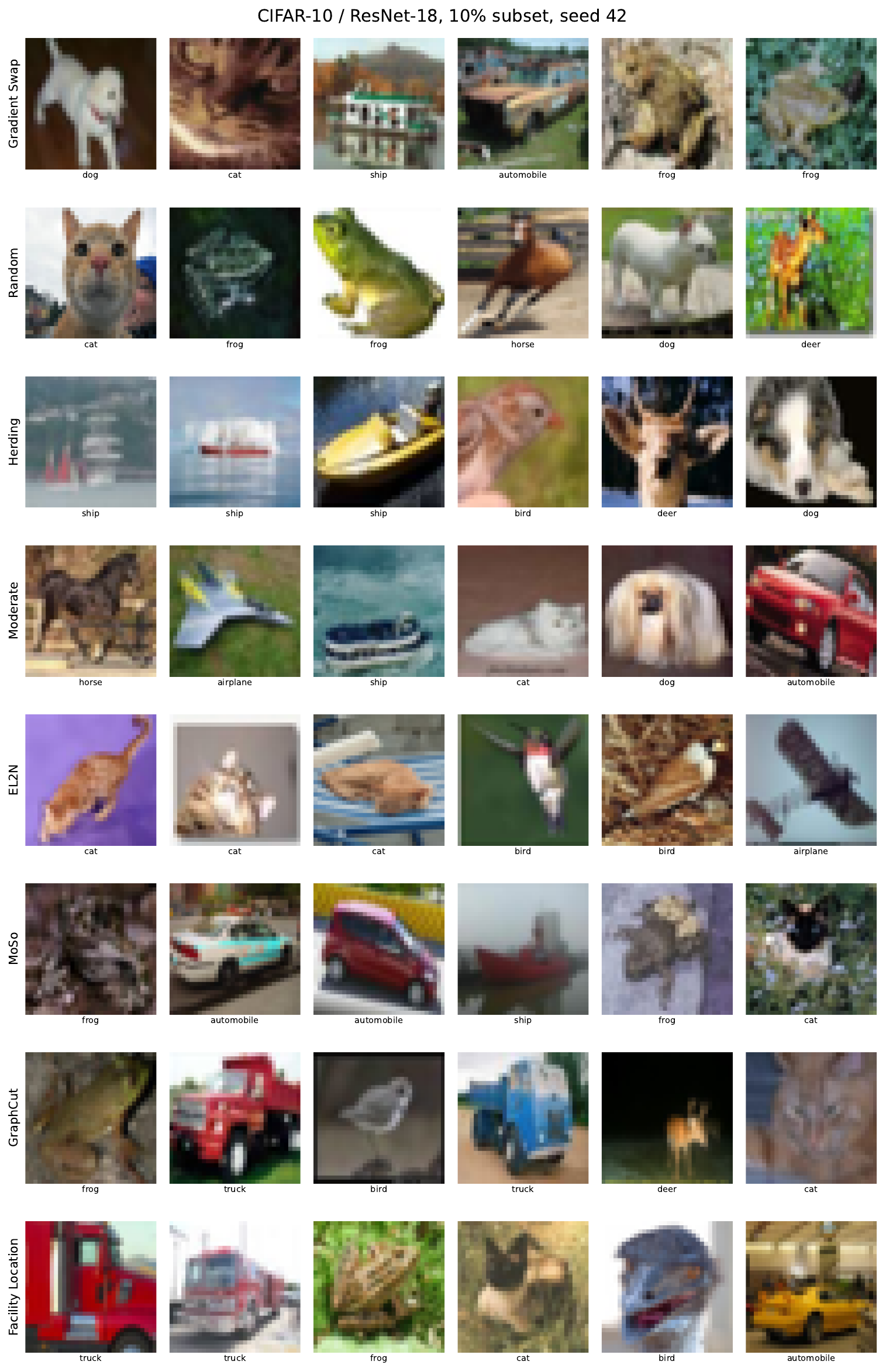}
\includegraphics[width=0.95\textwidth]{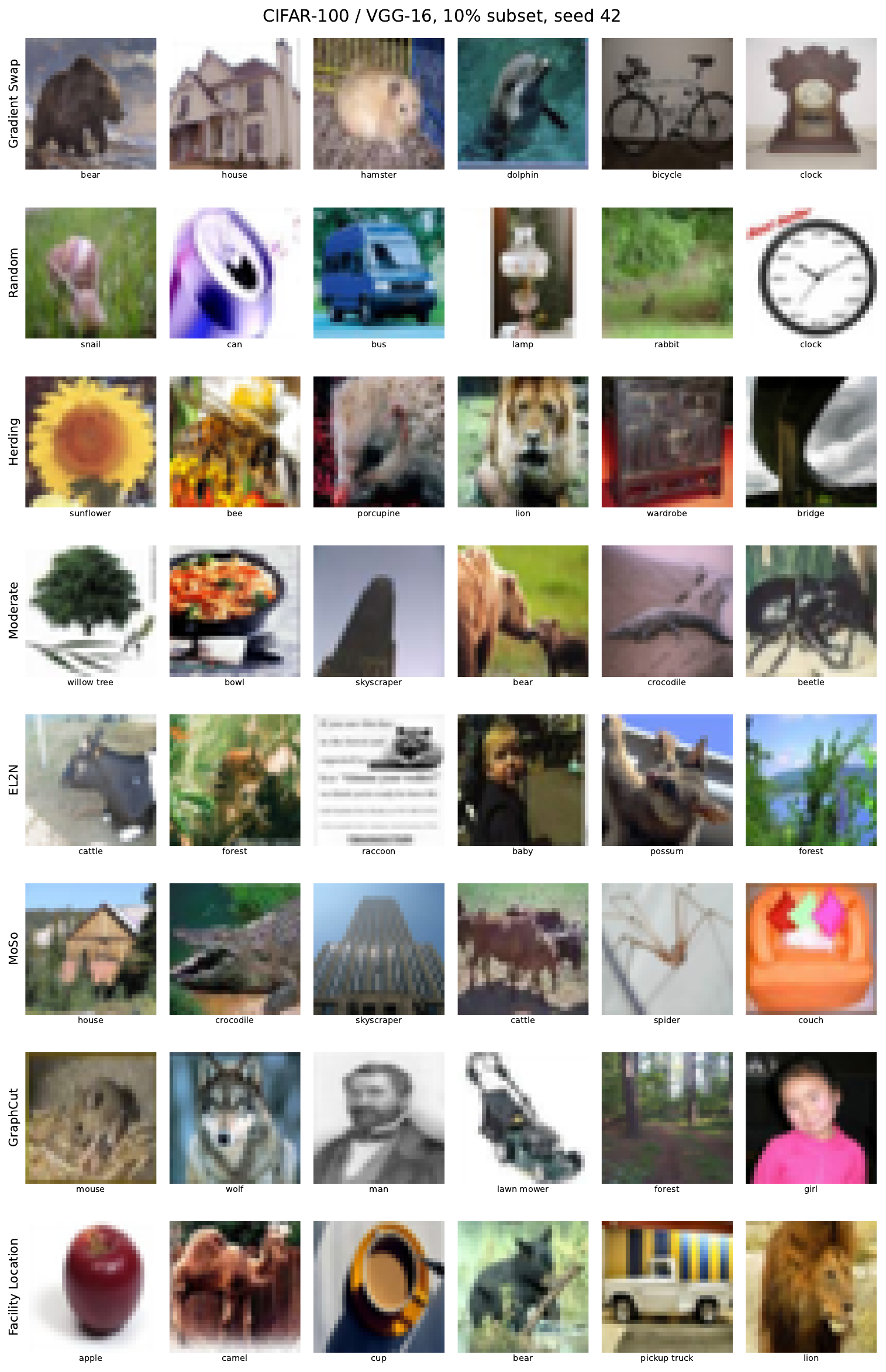}
\includegraphics[width=0.95\textwidth]{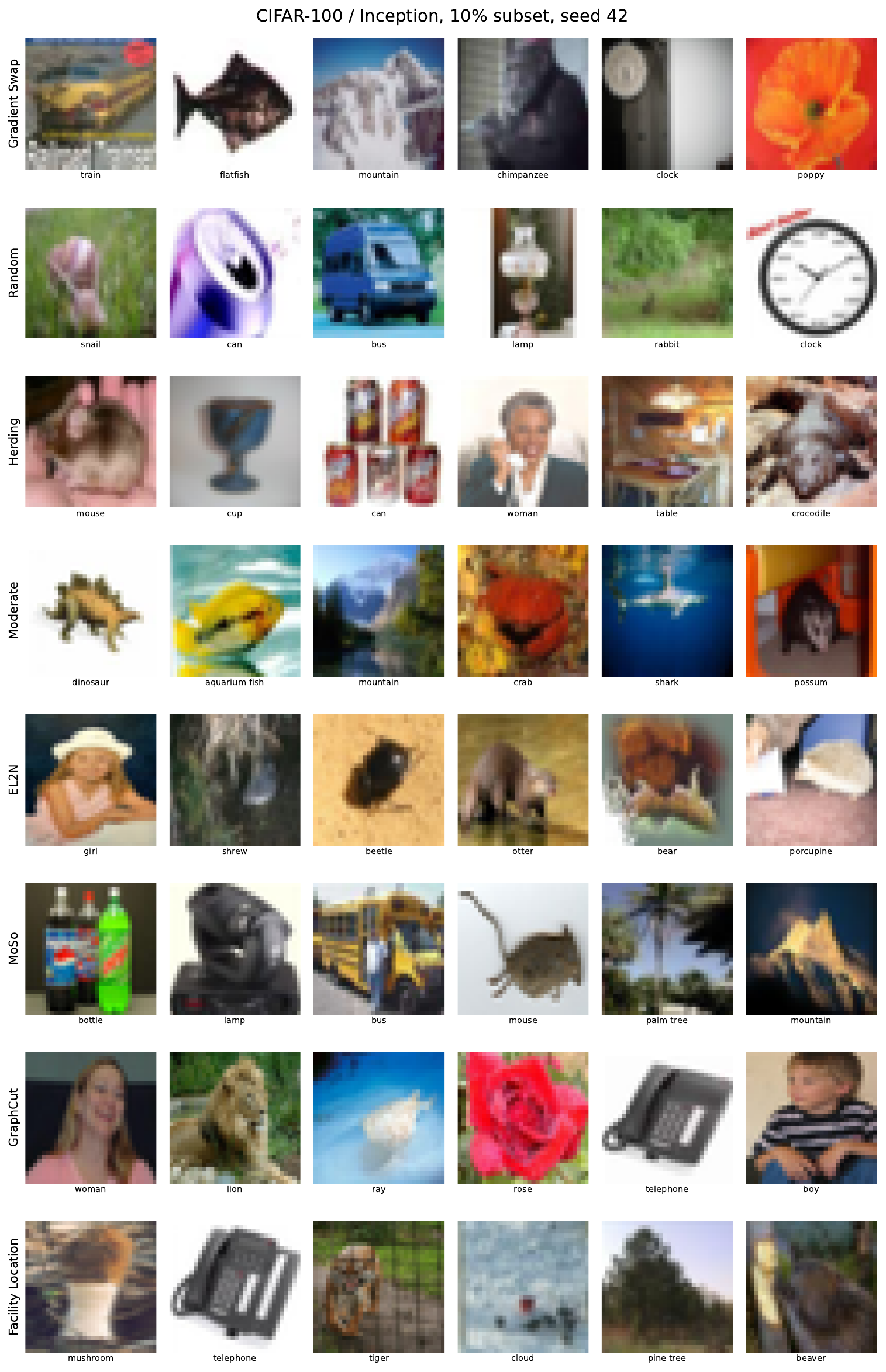}
\includegraphics[width=0.95\textwidth]{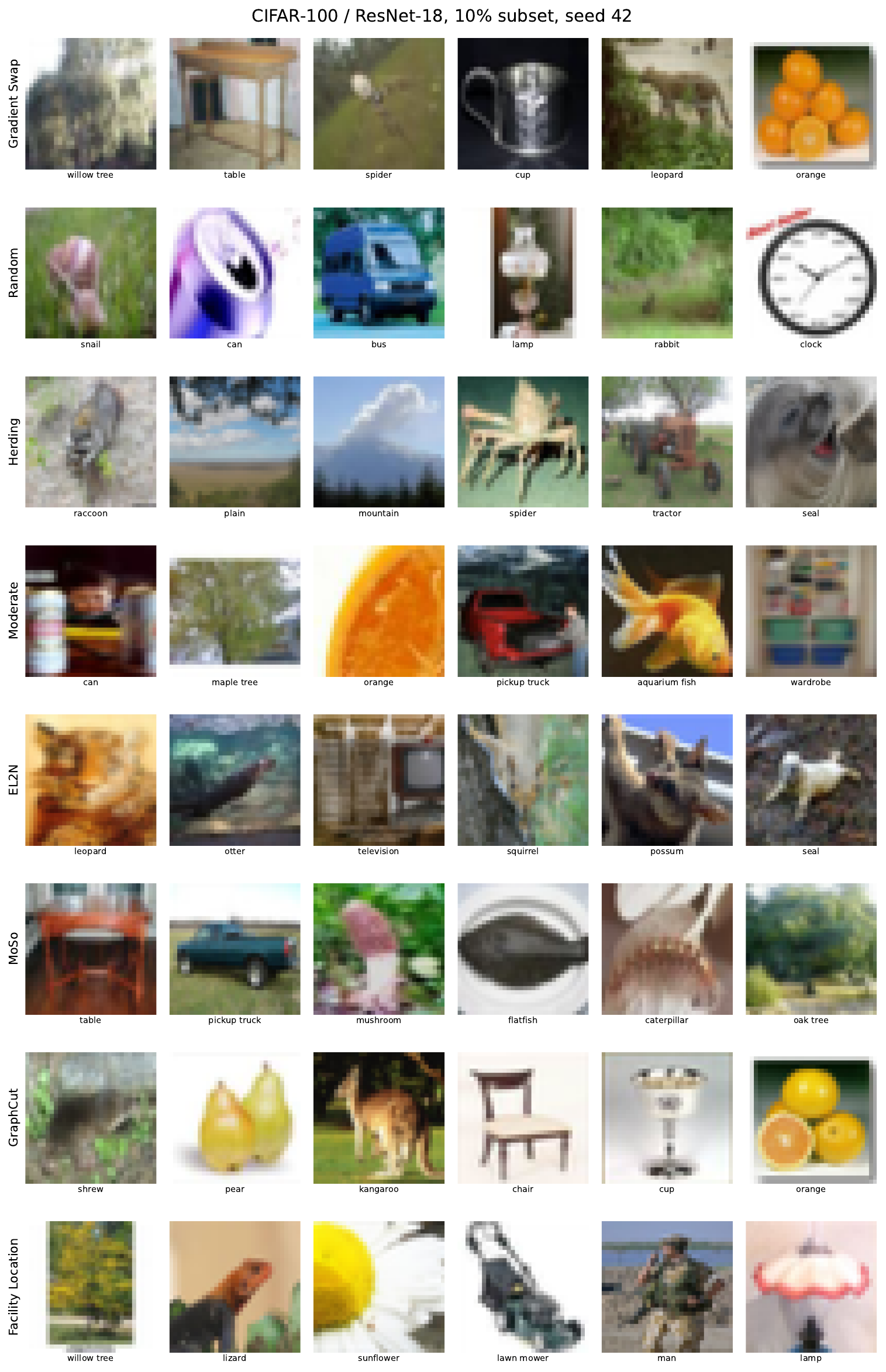}

%% file: arxiv_main.bbl
\begin{thebibliography}{}

\bibitem[Attias et~al., 2024]{attias_arxiv}
Attias, I., Dziugaite, G.~K., Haghifam, M., Livni, R., and Roy, D.~M. (2024).
\newblock Information complexity of stochastic convex optimization: Applications to generalization and memorization.
\newblock {\em ICML 2024}.

\bibitem[Brown et~al., 2021]{brown2021memorization}
Brown, G., Bun, M., Feldman, V., Smith, A., and Talwar, K. (2021).
\newblock When is memorization of irrelevant training data necessary for high-accuracy learning?
\newblock In {\em Proceedings of the 53rd annual ACM SIGACT symposium on theory of computing}, pages 123--132.

\bibitem[Chien et~al., 2024]{chien_langevin_2024}
Chien, E., Wang, H., Chen, Z., and Li, P. (2024).
\newblock Langevin {Unlearning}: {A} {New} {Perspective} of {Noisy} {Gradient} {Descent} for {Machine} {Unlearning}.
\newblock \_eprint: 2401.10371.

\bibitem[Cohen-Addad et~al., 2026]{cohen2026active}
Cohen-Addad, V., Kunapuli, S., Mirrokni, V., Nikdan, M., Woodruff, D.~P., and Zhou, S. (2026).
\newblock Active learning with low-rank structure for data selection.
\newblock {\em arXiv preprint arXiv:2606.16045}.

\bibitem[Coleman et~al., 2019]{coleman2019selection}
Coleman, C., Yeh, C., Mussmann, S., Mirzasoleiman, B., Bailis, P., Liang, P., Leskovec, J., and Zaharia, M. (2019).
\newblock Selection via proxy: Efficient data selection for deep learning.
\newblock {\em arXiv preprint arXiv:1906.11829}.

\bibitem[El~Halabi et~al., 2023]{el2023difference}
El~Halabi, M., Orfanides, G., and Hoheisel, T. (2023).
\newblock Difference of submodular minimization via dc programming.
\newblock In {\em International Conference on Machine Learning}, pages 9172--9201. PMLR.

\bibitem[Feldman, 2020]{feldman_does_2020}
Feldman, V. (2020).
\newblock Does learning require memorization? a short tale about a long tail.
\newblock In {\em Proceedings of the 52nd {Annual} {ACM} {SIGACT} {Symposium} on {Theory} of {Computing}}, pages 954--959, Chicago IL USA. ACM.

\bibitem[Feldman and Zhang, 2020]{feldman_what_2020}
Feldman, V. and Zhang, C. (2020).
\newblock What {Neural} {Networks} {Memorize} and {Why}: {Discovering} the {Long} {Tail} via {Influence} {Estimation}.
\newblock In {\em Advances in {Neural} {Information} {Processing} {Systems}}, volume~33, pages 2881--2891. Curran Associates, Inc.

\bibitem[Guo et~al., 2022]{guo2022deepcorecomprehensivelibrarycoreset}
Guo, C., Zhao, B., and Bai, Y. (2022).
\newblock Deepcore: A comprehensive library for coreset selection in deep learning.

\bibitem[Haghifam et~al., 2020]{haghifam_sharpened_2020}
Haghifam, M., Negrea, J., Khisti, A., Roy, D.~M., and Dziugaite, G.~K. (2020).
\newblock Sharpened {Generalization} {Bounds} based on {Conditional} {Mutual} {Information} and an {Application} to {Noisy}, {Iterative} {Algorithms}.
\newblock In {\em Advances in {Neural} {Information} {Processing} {Systems}}, volume~33, pages 9925--9935. Curran Associates, Inc.

\bibitem[Hanneke et~al., 2025]{pmlr-v291-hanneke25a}
Hanneke, S., Moran, S., Shlimovich, A., and Yehudayoff, A. (2025).
\newblock Data selection for erms.
\newblock In Haghtalab, N. and Moitra, A., editors, {\em Proceedings of Thirty Eighth Conference on Learning Theory}, volume 291 of {\em Proceedings of Machine Learning Research}, pages 2634--2665. PMLR.

\bibitem[He et~al., 2016]{he2016deep}
He, K., Zhang, X., Ren, S., and Sun, J. (2016).
\newblock Deep residual learning for image recognition.
\newblock In {\em Proceedings of the IEEE conference on computer vision and pattern recognition}, pages 770--778.

\bibitem[Inane et~al., 2026]{inane2026unlearningasymmetricsourcesimproved}
Inane, A.~M., Quirion, V., Dziugaite, G.~K., and Mitliagkas, I. (2026).
\newblock Unlearning with asymmetric sources: Improved unlearning-utility trade-off with public data.

\bibitem[Iyer et~al., 2021]{submodinformation}
Iyer, R., Khargoankar, N., Bilmes, J., and Asanani, H. (2021).
\newblock Submodular combinatorial information measures with applications in machine learning.
\newblock In Feldman, V., Ligett, K., and Sabato, S., editors, {\em Proceedings of the 32nd International Conference on Algorithmic Learning Theory}, volume 132 of {\em Proceedings of Machine Learning Research}, pages 722--754. PMLR.

\bibitem[Johnson et~al., 1984]{johnson1984extensions}
Johnson, W.~B., Lindenstrauss, J., et~al. (1984).
\newblock Extensions of lipschitz mappings into a hilbert space.
\newblock {\em Contemporary mathematics}, 26(189-206):1.

\bibitem[Killamsetty et~al., 2021a]{killamsetty_grad-match_2021}
Killamsetty, K., Sivasubramanian, D., Ramakrishnan, G., De, A., and Iyer, R. (2021a).
\newblock {GRAD}-{MATCH}: {Gradient} {Matching} based {Data} {Subset} {Selection} for {Efficient} {Deep} {Model} {Training}.
\newblock arXiv:2103.00123 [cs].

\bibitem[Killamsetty et~al., 2021b]{killamsetty2021glistergeneralizationbaseddata}
Killamsetty, K., Sivasubramanian, D., Ramakrishnan, G., and Iyer, R. (2021b).
\newblock Glister: Generalization based data subset selection for efficient and robust learning.

\bibitem[Kolossov et~al., 2023]{kolossov_towards_2023}
Kolossov, G., Montanari, A., and Tandon, P. (2023).
\newblock Towards a statistical theory of data selection under weak supervision.
\newblock arXiv:2309.14563 [stat.ML].

\bibitem[Krizhevsky, 2009]{alex2009learning}
Krizhevsky, A. (2009).
\newblock Learning multiple layers of features from tiny images.
\newblock {\em https://www. cs. toronto. edu/kriz/learning-features-2009-TR. pdf}.

\bibitem[Mirzasoleiman et~al., 2020]{mirzasoleiman_coresets_2020}
Mirzasoleiman, B., Bilmes, J., and Leskovec, J. (2020).
\newblock Coresets for {Data}-efficient {Training} of {Machine} {Learning} {Models}.
\newblock arXiv:1906.01827 [cs].

\bibitem[Moser et~al., 2025]{moser_coreset_2025}
Moser, B.~B., Shanbhag, A.~S., Frolov, S., Raue, F., Folz, J., and Dengel, A. (2025).
\newblock A {Coreset} {Selection} of {Coreset} {Selection} {Literature}: {Introduction} and {Recent} {Advances}.
\newblock Version Number: 2.

\bibitem[Narasimhan and Bilmes, 2012]{narasimhan2012submodularsupermodularprocedureapplicationsdiscriminative}
Narasimhan, M. and Bilmes, J.~A. (2012).
\newblock A submodular-supermodular procedure with applications to discriminative structure learning.

\bibitem[Negrea et~al., 2019]{negrea_information-theoretic_2019}
Negrea, J., Haghifam, M., Dziugaite, G.~K., Khisti, A., and Roy, D.~M. (2019).
\newblock Information-theoretic generalization bounds for {SGLD} via data-dependent estimates.
\newblock {\em Advances in Neural Information Processing Systems}, 32.

\bibitem[Nikdan et~al., 2026]{nikdan2026efficient}
Nikdan, M., Cohen-Addad, V., Alistarh, D., and Mirrokni, V. (2026).
\newblock Efficient data selection at scale via influence distillation.
\newblock {\em Advances in Neural Information Processing Systems}, 38:143229--143265.

\bibitem[Paul et~al., 2021]{paul_deep_2021}
Paul, M., Ganguli, S., and Dziugaite, G.~K. (2021).
\newblock Deep {Learning} on a {Data} {Diet}: {Finding} {Important} {Examples} {Early} in {Training}.
\newblock In {\em Advances in {Neural} {Information} {Processing} {Systems}}, volume~34, pages 20596--20607. Curran Associates, Inc.

\bibitem[Pia et~al., 2014]{delpia2014mixedintegerquadraticprogrammingnp}
Pia, A.~D., Dey, S.~S., and Molinaro, M. (2014).
\newblock Mixed-integer quadratic programming is in np.

\bibitem[Pruthi et~al., 2020]{pruthi_estimating_2020}
Pruthi, G., Liu, F., Sundararajan, M., and Kale, S. (2020).
\newblock Estimating {Training} {Data} {Influence} by {Tracing} {Gradient} {Descent}.
\newblock arXiv:2002.08484 [cs.LG].

\bibitem[Shalev-Shwartz and Ben-David, 2014]{shalev2014understanding}
Shalev-Shwartz, S. and Ben-David, S. (2014).
\newblock {\em Understanding machine learning: From theory to algorithms}.
\newblock Cambridge university press.

\bibitem[Simonyan and Zisserman, 2015]{simonyan2015deepconvolutionalnetworkslargescale}
Simonyan, K. and Zisserman, A. (2015).
\newblock Very deep convolutional networks for large-scale image recognition.

\bibitem[Sorscher et~al., 2023]{sorscher2023neuralscalinglawsbeating}
Sorscher, B., Geirhos, R., Shekhar, S., Ganguli, S., and Morcos, A.~S. (2023).
\newblock Beyond neural scaling laws: beating power law scaling via data pruning.

\bibitem[Szegedy et~al., 2014]{szegedy2014goingdeeperconvolutions}
Szegedy, C., Liu, W., Jia, Y., Sermanet, P., Reed, S., Anguelov, D., Erhan, D., Vanhoucke, V., and Rabinovich, A. (2014).
\newblock Going deeper with convolutions.

\bibitem[Tan et~al., 2023]{tan2023data}
Tan, H., Wu, S., Du, F., Chen, Y., Wang, Z., Wang, F., and Qi, X. (2023).
\newblock Data pruning via moving-one-sample-out.
\newblock {\em Advances in neural information processing systems}, 36:18251--18262.

\bibitem[Toneva et~al., 2019]{toneva2019empiricalstudyexampleforgetting}
Toneva, M., Sordoni, A., des Combes, R.~T., Trischler, A., Bengio, Y., and Gordon, G.~J. (2019).
\newblock An empirical study of example forgetting during deep neural network learning.

\bibitem[Wei et~al., 2015]{pmlr-v37-wei15}
Wei, K., Iyer, R., and Bilmes, J. (2015).
\newblock Submodularity in data subset selection and active learning.
\newblock In Bach, F. and Blei, D., editors, {\em Proceedings of the 32nd International Conference on Machine Learning}, volume~37 of {\em Proceedings of Machine Learning Research}, pages 1954--1963, Lille, France. PMLR.

\bibitem[Welling, 2009]{welling2009herding}
Welling, M. (2009).
\newblock Herding dynamical weights to learn.
\newblock In {\em Proceedings of the 26th annual international conference on machine learning}, pages 1121--1128.

\bibitem[Welling and Teh, 2011]{wellingteh}
Welling, M. and Teh, Y.~W. (2011).
\newblock Bayesian learning via stochastic gradient langevin dynamics.
\newblock In {\em Proceedings of the 28th International Conference on International Conference on Machine Learning}, ICML'11, page 681–688, Madison, WI, USA. Omnipress.

\bibitem[Xia et~al., 2024]{xia2024less}
Xia, M., Malladi, S., Gururangan, S., Arora, S., and Chen, D. (2024).
\newblock Less: Selecting influential data for targeted instruction tuning.
\newblock {\em arXiv preprint arXiv:2402.04333}.

\bibitem[Xia et~al., 2022]{xia2022moderate}
Xia, X., Liu, J., Yu, J., Shen, X., Han, B., and Liu, T. (2022).
\newblock Moderate coreset: A universal method of data selection for real-world data-efficient deep learning.
\newblock In {\em The Eleventh International Conference on Learning Representations}.

\bibitem[Ye et~al., 2024]{ye_leave-one-out_2024}
Ye, J., Borovykh, A., Hayou, S., and Shokri, R. (2024).
\newblock Leave-one-out {Distinguishability} in {Machine} {Learning}.
\newblock arXiv:2309.17310 [cs].

\bibitem[Zha et~al., 2025]{zha2025data}
Zha, D., Bhat, Z.~P., Lai, K.-H., Yang, F., Jiang, Z., Zhong, S., and Hu, X. (2025).
\newblock Data-centric artificial intelligence: A survey.
\newblock {\em ACM Computing Surveys}, 57(5):1--42.

\end{thebibliography}
